\documentclass[pdflatex,sn-vancouver-ay]{sn-jnl}% Vancouver Author Year Reference Style
\usepackage{graphicx}%
\usepackage{multirow}%
\usepackage{amsmath,amssymb,amsfonts}%
\usepackage{amsthm}%
\usepackage{mathrsfs}%
\usepackage[title]{appendix}%
\usepackage{xcolor}%
\usepackage{textcomp}%
\usepackage{manyfoot}%
\usepackage{booktabs}%
\usepackage{placeins}
\usepackage{algorithm}%
\usepackage{algorithmicx}%
\usepackage{algpseudocode}%
\usepackage{listings}%
\usepackage{amssymb}
\usepackage{amsmath}
\usepackage{graphicx}
\usepackage{booktabs}
\usepackage{multirow}
\usepackage{xcolor}
\usepackage{hyperref}
\usepackage{float}
\usepackage{caption}
\usepackage{subcaption}
\usepackage{algorithm}
\usepackage{algpseudocode}
\usepackage{float}
\theoremstyle{thmstyleone}%
\theoremstyle{thmstyletwo}%

\theoremstyle{thmstylethree}%

\begin{document}

\title[Article Title]{Enhancing Cross-Grade Generalization in Banana Variety Classification via Hybrid CNN-Transformers and Knowledge Self-Distillation}

%%=============================================================%%
%% GivenName	-> \fnm{Joergen W.}
%% Particle	-> \spfx{van der} -> surname prefix
%% FamilyName	-> \sur{Ploeg}
%% Suffix	-> \sfx{IV}
%% \author*[1,2]{\fnm{Joergen W.} \spfx{van der} \sur{Ploeg} 
%%  \sfx{IV}}\email{iauthor@gmail.com}
%%=============================================================%%

\author[1]{\fnm{Khush} \sur{Kataruka}\email{u23ai112@coed.svnit.ac.in}}

\author[1]{\fnm{Mohammad} \sur{Muzammila}\email{u23ai105@coed.svnit.ac.in}}
%\equalcont{These authors contributed equally to this work.}

\author[2]{\fnm{Zahid} \sur{Akhtar}\email{akhtarz@sunypoly.edu}}

\author[3]{\fnm{Raghavendra} \sur{Ramachandra}\email{raghavendra.ramachandra@ntnu.no}}

\author*[1]{\fnm{Praveen Kumar} \sur{Chandaliya}\email{pkc@aid.svnit.ac.in}}
%\equalcont{These authors contributed equally to this work.}

\affil[1]{\orgdiv{Department of Artificial Intelligence},
	\orgname{Sardar Vallabhbhai National Institute of Technology},
	\city{Surat},
	\state{Gujarat},
	\country{India}}

\affil[2]{\orgdiv{Electrical and Computer Engineering Department},
	\orgname{State University of New York Polytechnic Institute},
	\city{Utica},
	\state{New York},
	\country{USA}}

\affil[3]{\orgdiv{Department of Information Security and Communication Technology},
	\orgname{Norwegian University of Science and Technology},
	\city{Gjovik},
	\country{Norway}}
%%==================================%%
%% Sample for unstructured abstract %%
%%==================================%%

\abstract{In recent years, automated fruit classification systems have become a more critical component in precision agriculture for efficient yield estimation, quality control, and supply chain optimization. However, they still face significant challenges under varying quality, fruit maturity, illumination, texture degradation, morphological attributes, domain shift, and post-harvest handling conditions. In this study, we address the challenging issue of cross-grade banana variety classification. This is a real-world setting where models that have been trained on specific quality grades must generalize to unseen grades to handle significant domain shift. Under a rigorous Leave-One-Grade-Out (LOGO) evaluation protocol, we present a thorough comparative analysis of convolutional neural networks (CNNs), Vision Transformers (ViTs), and dual-stream hybrid architectures using a large scale dataset of 31,678 banana images with three varieties and three quality grades. Specifically, to address the drawbacks of single-paradigm frameworks, we present a dual-stream Swin ResNet hybrid framework enhanced with KL-divergence-based self-distillation to align probabilistic feature representations across the CNN and Transformer streams. Comprehensive experimental analyzes show that the presented hybrid framework with KL-divergence based self-distillation achieves better overall performance, i.e., it attains a mean accuracy of $81.25\%$ and significantly surpasses CNN and Transformer baseline models. Qualitative analysis further confirm that the hybrid framework in this study improves discriminative feature transferability, improves attention localization, and stabilizes optimization. Overall, the experimental results  demonstrate how well hybrid distillation techniques can lead to domain robust agricultural systems and can be considered a promising direction for robust agricultural vision frameworks in the real world.}


%%\pacs[JEL Classification]{D8, H51}

%%\pacs[MSC Classification]{35A01, 65L10, 65L12, 65L20, 65L70}

\maketitle

\section{Introduction}
\label{sec:intro}

Automated fruit variety classification is becoming increasingly vital in modern farming, as it supports large-scale yield estimation, quality control, and a smooth supply chain. But variations in growing regions, inconsistent cultivation practices, improper handling during harvest and distribution, inadequate disease and pest management make bananas susceptible to quality issues like bruising, peel discoloration, premature ripening, fungal lesions, splitting, shape deformities, and uneven coloration \cite{ALDAIRI2023177}. These imperfections significantly diminish retail value and marketability, resulting in substantial economic losses for producers and supply chain stakeholders \cite{fernando2019quality}. At present, the prevailing grading approach still depends heavily on manual inspection that suffers from low throughput, high labor costs and variable assessment criteria influenced by human subjectivity. Although mechanical sorting systems can enhance grading accuracy, they remain incapable of effectively assessing surface imperfections like bruising, spotting, or peel discoloration. Also, physical contact during sorting often results in secondary damage.
These challenges and bananas being one of the world's most vital staple crops (e.g., global production exceeding 114 million tonnes annually and serving as a primary food source for over 400 million people worldwide \cite{app15063195,Zhang2026}) have motivated extensive research into automated classification systems using machine learning and computer vision techniques \cite{ZHANG2024100733}. 

Machine learning and computer vision-based automated classification systems offer substantial benefits, such as increased processing speed, consistent quality evaluation, reduced labour costs, and enhanced traceability throughout the supply chain \cite{ZHANG2024100733}. Early techniques mainly employed hand-crafted features (e.g., color histograms, texture descriptors) combined with classical machine learning algorithms such as Support Vector Machines (SVMs) and Random Forests. For example, the method in \cite{s19132910,SANTOSPEREIRA201876} extracted peel color, local texture, and local shape features from different banana regions using local binary pattern (LBP) and classifying them using (Naive Bayes) NB, Linear Discriminant Analysis (LDA), and SVM algorithms \cite{GHAZAL2021106267}. While \cite{FU2019105057} presented a multi-scale method that used VM and AdaBoost algorithms with fused LBP (local binary pattern) and HOG (histogram of oriented gradients) features. Nonetheless, these methods demonstrated limited generalization capability (i.e., failing to capture the complex hierarchical representations), especifically when confronting variations in lighting conditions, camera angles, and fruit quality grades.

With recent advances in computer vision and deep learning, intelligent inspection and sorting systems have emerged as a critical solution to overcoming these limitations. The deep learning algorithms have catalyzed a paradigm shift in agricultural automation, including fruit quality assessment and variety classification. In particular, Convolutional Neural Networks (CNNs) have revolutionized computer vision tasks by enabling automatic feature extraction directly from raw image data \cite{li2021survey}. CNNs have demonstrated exceptional success across various agricultural applications, including crop disease detection, fruit ripeness assessment, and variety classification \cite{wallelign2018soybean,olisah2024convolutional,Gurjar2026, ponce2019olive}. For instance, the study in \cite{SHEIKH202518,XU2026111778} employed deep convolutional features learned from banana images using a transfer learning framework. Specifically, a fine-tuned deep learning classifier was used to distinguish banana varieties based on shape, color, and texture cues captured in the learned feature maps. While, the work in \cite{SHUPRAJHAA2023112410} proposed a CNN-XGBoost hybrid model for classifying banana ripening stages, where a convolutional neural network extracts deep visual features such as color, texture, and shape, and XGBoost performs the classification. Moreover, the proposed system used Linear Discriminant Analysis (LDA) for dimensionality reduction, thereby enabling effective training with smaller datasets. 

Despite remarkable advances in deep learning for automated classification of fruit and banana varieties, several crucial challenges remain inadequately addressed \cite{TOO2019272,MEHDIPOUR2026200617,SARODE2026111669}. Most current models (e.g., CNNs, Vision Transformers (ViTs), or hybrid architectures) perform well on in-distribution data but struggle to generalize across domain shifts commonly encountered in real-world banana production and distribution as well as other agriculture settings. These shifts arise from variations in quality grades, maturity levels, lighting conditions, and camera setups, leading to substantial degradation in model performance when confronted with unseen distributions of visual or quality-grade features. CNNs, while effective at capturing local texture and shape information, often fail to model global structural relationships critical for distinguishing visually similar varieties. Conversely, ViTs capture long-range dependencies and global context but typically require large datasets and may underutilize local texture cues that remain discriminative under quality degradation. Hybrid approaches that combine the strengths of both paradigms show promise, yet their efficacy for cross-grade and cross-domain agricultural classification remains largely unexplored. Consequently, ensuring robust, reliable performance across diverse fruit quality conditions remains an open and pressing research problem in precision banana agriculture. All in all, not only do existing methods struggle to generalize when confronted with unseen quality grades (particularly in settings where inter-class differences are subtle and intra-class variability is high), but the effectiveness of hybrid architectures that combine complementary strengths for cross-grade classification remains underexplored.

As deep learning and vision-based systems are increasingly being deployed in banana orchards, packing houses, and markets, robustness to variations in grades, lighting, maturity, and surface quality is as crucial as classification accuracy. Thus, in this work, we focus on cross-grade banana variety classification, a challenging yet practically relevant task. To address the limitations of single-paradigm models, we propose a dual-stream CNN-Transformer hybrid architecture augmented with KL-divergence-based distillation. By encouraging alignment between the predictive distributions of a Swin Transformer (teacher) and a ResNet-based CNN (student), the model learns grade-invariant representations that are more robust to domain shift.
The primary novelty of this work lies not only in the proposed CNN-Transformer architecture, but also in establishing cross-grade generalisability as a distinct and practically meaningful evaluation problem for banana variety classification.  This paper makes the following key contributions:
\begin{itemize}
	\item  We establish a Leave-One-Grade-Out (LOGO) cross-validation framework, in which each grade is completely held out during testing, providing a more realistic assessment of model robustness to unseen grades than conventional random splits.
	\item  We systematically evaluate CNNs, Vision Transformers, and hybrid architectures under the LOGO protocol, thereby offering one of the first large-scale studies of banana variety classification under a rigorous cross-grade evaluation protocol.
	\item  We introduce a hybrid model integrating a Swin Transformer and ResNet backbone to leverage feature fusion to improve cross-grade generalization and outperform pure CNN or Transformer baselines.
	\item  We propose a self-distillation mechanism that aligns probability distributions between architectural streams, significantly improving robustness on challenging domains and accelerating convergence.
	
	\item  Through attention-based visualizations (Grad-CAM and Score-CAM), inter-stream alignment analysis, and training dynamics studies, we offer in-depth insights into how hybrid representations and knowledge distillation jointly mitigate domain shift effects.
\end{itemize}
To the best of our knowledge, we present the first systematic investigation of banana variety classification under a rigorous cross-grade generalisation setting. The paper is organized as follows. Section \ref{sec:Relatedwork} reviews related work, and Section \ref{sec:dataset} describes the dataset. Section \ref{sec:proposedframework} presents the problem formulation, our proposed hybrid architecture, and KL-divergence-based feature alignment. Section \ref{sec:Experiments} details the experimental analysis, including quantitative, interpretability, and qualitative results. Section \ref{sec:conclusion} presents conclusions with future research directions.

\section{Related Work}
\label{sec:Relatedwork}

The utilization of computer vision (CV) and machine learning (ML) to horticultural variety classification has evolved considerably over the past two decades. CV and ML have been widely applied to fruit detection and classification tasks, including apple, mango, citrus, and banana varieties \cite{TOO2019272}. In particular, most initial works in this field typically involved manual feature engineering, i.e., extracting local and texture characteristics, e.g., color histograms, image texture descriptors and shape features \cite{liakos2018machine}. For example, the study in \cite{8257854} extracted scale-invariant shape features and used a Bayesian-based classifier for banana cultivar classification. Other works include the use of geometrical, shape, and color features with discriminant analysis for corn variety recognition \cite{CHEN2010S48}, principal component analysis for basmati rice classification \cite{kambo2014classification}, color statistical and texture features for tomato grading \cite{ARAKERI2016426}, and morpho-colorimetric descriptors and LDA to classify plum varieties \cite{SARIGU201725}. Such traditional ML based systems achievied moderate success under controlled settings but demonstrated lower robustness to environmental variations and required significant domain expertise for feature design.

The rise of deep learning reduced dependence on hand-crafted features and traditional machine learning classifiers. Researchers have successfully employed deep learning architectures for various fruit classification applications and achieved higher accuracy than traditional techniques. In particular, most studies have used CNNs as the dominant approach in agricultural image analysis \cite{zangana2024advancements}. Models such as AlexNet, VGG, ResNet, and EfficientNet have achieved strong performance in fruit defect detection, classification, and grading. For instance, authors in \cite{al2021cnn} devised a CNN-based framework for strain detection and quality identification of bananas. The work in \cite{dinata2021classification} utilised a CNN-based classification framework for strawberry plant disease detection. Recently, Vision Transformers (ViTs) have provided an alternative to CNNs by modeling self-attention mechanisms to model global dependencies in images and videos \cite{mehdipour2025vision}. Research studies have shown that ViTs can rival or exceed CNNs on some agriculture tasks. For instance, the study in \cite{10593120} classified banana plant leaves based on nutrient deficiency using a vision transformer. The problem of fine-grained, multi-defect classification for green plums was addressed using an improved vision transformer in \cite{su2023multiple}. Nonetheless, Transformer-based frameworks usually require large-scale data and fail to assess cross-domain robustness.

Hybrid CNN-Transformer methods aim to integrate the complementary strengths of CNNs, which efficiently capture local texture features, and transformers, which model long-range dependencies \cite{khan2023survey}. Various fusion schemes, including early, late, and attention-based, have been developed to combine these paradigms. A hybrid CNN?Transformer model was proposed in \cite{li2020cnn} for multitemporal crop classification, where multisensor data are aligned and Transformer encoders are employed to capture temporal correlations. Furthermore, a framework that integrates local and global feature representations using a CNN?Transformer architecture was introduced in \cite{10776810}. Similarly, Knowledge-distillation (KD), which was traditionally utilized for model compression, has been extended to feature alignment and cross-architecture learning \cite{moslemi2024survey}. KL-divergence has been widely employed to align probability distributions between teacher and student networks. The systems in \cite{10826582} used KD to train a lightweight MobileNetV3 with a ViT as the teacher for green plum defect classification. Despite the remarkable success of deep learning, including hybrid CNN?Transformer models and KD in other vision tasks, these approaches remain largely unexplored for agricultural generalization classification under cross-domain conditions. Thus, our work seeks to address this gap.

\section{Dataset}\label{sec:dataset}
This study used a comprehensive dataset consisting of 31,678 high-resolution banana images collected across three commercial varieties (i.e., Pachbale, Rasbale, and Elakki-bale). Such varieties are mainly grown in South Asian agricultural regions \cite{GURU2025111478}. These varieties exhibit subtle morphological differences that challenge both automated classification systems and human experts. Three quality grades (e.g., Grades 1: heavily blemished, 2: moderate defects, and 3: premium quality) are utilized to represent each variety. These grades do not reflect variety-specific attributes but rather differences in ripeness, physical condition and marketability. Overall, the dataset is evenly balanced across varieties and grades (as detailed in Tables~\ref{tab:dataset_overview} and \ref{tab:grade_distribution}), ensuring that no single class dominates the training and evaluation.

As shown in Table \ref{tab:grade_distribution}, each quality grade comprises approximately 10,000 samples, resulting in a relatively uniform distribution across grades. This even distribution ensures that accuracy differences largely come from domain shift rather than class imbalance. Grade 1 samples (i.e., 32.87\%) show the highest visual noise and variability, and Grade 3 samples (i.e., 32.94\%) are the most visually consistent and represent the easiest domain.

\begin{table*}[h]
	\centering
	\caption{Overall Dataset Statistics.}
	\label{tab:dataset_overview}
	\begin{tabular}{lrr}
		\toprule
		\textbf{Variety} & \textbf{Count} & \textbf{Percentage} \\
		\midrule
		Pachbale & 11,512 & 36.34\% \\
		Rasbale & 10,112 & 31.92\% \\
		Elakki-bale & 10,054 & 31.74\% \\
		\midrule
		\textbf{Total} & \textbf{31,678} & \textbf{100.00\%} \\
		\bottomrule
	\end{tabular}
\end{table*}

\begin{table*}[h]
	\centering
	\caption{Variety Distribution Across Quality Grades}
	\label{tab:grade_distribution}
	\begin{tabular}{ccrrr}
		\toprule
		\textbf{Grade} & \textbf{Variety} & \textbf{Count} & \textbf{Percentage} & \textbf{Grade Total} \\
		\midrule
		
		\multirow{3}{*}{\begin{tabular}{c} Grade 1 \\ (Heavily blemished) \end{tabular}} 
		& Pachbale      & 3,807 & 36.56\% & \multirow{3}{*}{10,412} \\
		& Rasbale       & 3,589 & 34.47\% &                         \\
		& Elakki-bale   & 3,016 & 28.97\% &                         \\
		\midrule
		
		\multirow{3}{*}{\begin{tabular}{c} Grade 2 \\ (Moderate defects) \end{tabular}} 
		& Pachbale      & 3,944 & 36.42\% & \multirow{3}{*}{10,830} \\
		& Elakki-bale   & 3,571 & 32.97\% &                         \\
		& Rasbale       & 3,315 & 30.61\% &                         \\
		\midrule
		
		\multirow{3}{*}{\begin{tabular}{c} Grade 3 \\ (Premium quality) \end{tabular}} 
		& Pachbale      & 3,761 & 36.04\% & \multirow{3}{*}{10,436} \\
		& Elakki-bale   & 3,467 & 33.22\% &                         \\
		& Rasbale       & 3,208 & 30.74\% &                         \\
		\bottomrule
	\end{tabular}
	\label{tab:table2}
\end{table*}

\section{Proposed Cross-Grade Banana Variety Classification Methodology}
\label{sec:proposedframework}

\subsection{Problem Formulation}

Under domain generalization, we formulate the cross-grade banana variety recognition as a multi-class classification problem. The objective is to accurately identify the banana variety samples that come from an unknown grade by acquiring samples that belong to specific quality grades. Let $X$ and $Y=\{\text{Pachbale}, \text{Rasbale}, \text{Elakki-bale}\}$ denote the input sample space and set of variety labels, respectively. The dataset $D$ is splited into three quality dependent domains, i.e., $D_1$, $D_2$, and $D_3$ matching to Grade~1, Grade~2, and Grade~3, respectively. More formally, let  $D=\{(x_i, y_i, g_i)\}_{i=1}^{N}$, where $x_i \in \mathbb{R}^{H \times W \times 3}$, $y_i \in Y$, and $g_i \in \{1,2,3\}$ indicate an RGB sample, variety label, and quality grade (domain), respectively.

The distribution shift across grades is the problem. The distributions of visual features $P(X \mid Y, g)$ vary with $g$ owing to differences in ripeness, defects, texture, and surface appearance. As a result, a model that has been trained on one grade might not naturally generalize to another. The aim is to learn a classifier $f_{\theta}: X \rightarrow Y$ utilizing training domains $G_{\text{train}} \subset \{1,2,3\}$ so that it functions effectively on an unseen domain $g_{\text{test}} \notin G_{\text{train}}$. Thus, maximizing predictive accuracy on the held-out $D_{\text{test}}$ is the goal, while only being able to access/train on samples from the $D_{\text{train}}$. In this study, the Leave-One-Grade-Out (LOGO) protocol is used for evaluation. In every run, one grade is held out for testing, and the remaining grades are employed for training: $G_{\text{train}} = \{1,2,3\} \setminus \{g_{\text{test}}\}$. For example, if the system is being testing on grade 2 then it was trained on grades 1 and 3. This formulation differs from standard supervised classification, where the training and test samples are assumed to come from the same distribution. The cross-grade scenarios give a more realistic picture of what actually happens in the real-world agriculture. The systems have to handle and remain reliable across a wide range of quality differences, including grades that were not part of the training set. This approach actually captures that challenge.

\subsection{Baselines Architectures}
In this study, to establish solid reference points, three popular vision paradigms (i.e., ResNet50 \cite{he2016deep}, EfficientNet-B1 \cite{tan2019efficientnet}, and Swin Transformer \cite{liu2021swin}) were evaluated as baselines. Under the same training conditions, all models were fine-tuned on the banana classification dataset after being initialized with ImageNet pre-trained weights. \emph{ResNet50} is the main CNN baseline. Its residual skip connections allow for stable deep network optimization and mitigate vanishing gradient problems. The model comprises four residual stages with gradually increasing channel dimensions, global average pooling, and a task-specific classifier. \emph{EfficientNet-B1} provides a parameter-efficient CNN comparison. It is based on compound scaling that balances input resolution, network width, and depth. Compared with conventional CNNs, the architecture offers robust trade-offs between accuracy and efficiency by using MBConv blocks with squeeze-and-excitation modules. \emph{Swin Transformer Base} portrays the transformer family. Unlike standard Vision Transformers, Swin uses shifted-window self-attention and hierarchical patch merging to capture long-range dependencies at a lower computational cost. The base configuration utilizes four stages with progressively larger embedding dimensions. It is worth noting that the models are initialized with ImageNet-V2 (for ResNet50 and EfficientNet-B1) and ImageNet-1K (for Swin Transformer) pretrained weights and fine-tuned on the banana dataset.

%We evaluate distinct architectures representing different paradigms in computer vision:

%\subsubsection{Baselines (CNN \& Transformer)}
%\begin{itemize}
%   \item \textbf{ResNet50}: 50-layer residual network (ImageNet V2 weights).
%   \item \textbf{EfficientNet-B1}: Compound-scaled efficient architecture (ImageNet V2 weights).
%  \item \textbf{Swin-Tiny}: Hierarchical Vision Transformer (ImageNet-1K weights).
%\end{itemize}

\subsection{Hybrid Teacher-Student Framework with KL-Divergence Distillation}

To overcome the drawbacks of uni-paradigm architectures under cross-grade domain shift, we formulate a unified and dual-stream teacher?student hybrid framework which combines a Swin Transformer-based global representation learner with a ResNet50-based local feature extractor. Namely, the developed framework integrates global contextual relationships via a transformer and local texture patterns via CNNs. The architecture comprises six main parts: a Swin-Tiny teacher network, a ResNet50 student network, a learnable feature adaptor, dual classification heads, logit-level fusion, and a joint supervision and distillation objective. Two variations of the designed architecture were created in order to improve generalization and encourage coherent learning between the streams, i.e., a standard hybrid mechanism trained with cross-entropy loss on fused outputs alone, and a KL-distillation hybrid mechanism that additionally incorporates a KL divergence alignment term between the two streams.

%We propose a dual-stream architecture that leverages both local texture features (via CNN) and global context (via Swin Transformer).

%\begin{itemize}
%   \item \textbf{Model Structure}: The hybrid architecture consists of two parallel backbones: a $\textbf{Swin Transformer} (Teacher) and a $\textbf{ResNet50} (Student). Features from both streams are fused to produce the final classification.
%   \item \textbf{Standard Hybrid}: This variant uses the exact same dual-stream structure but is trained using standard Cross-Entropy loss on the fused output only. It does not employ the auxiliary KL-divergence alignment term.
%  \item \textbf{KL-Distillation Hybrid}: This variant uses the same structure but incorporates a KL Divergence loss term to enforce feature distribution alignment between the Swin and CNN branches.
%\end{itemize}

%\subsubsection{Hybrid Dual Stream Design:} 
In the proposed architecture, the same input sample is processed by two parallel backbones. i.e., teacher branch (Swin Transformer as global representation learner) and student branch (ResNet50 as local feature extractor). Given an input image $x \in \mathbb{R}^{224 \times 224 \times 3}$ and ground-truth label $y \in \{1,\dots,C\}$, the goal is to learn a robust predictive function by combining complementary high-level representations extracted by two branches separately. In fact, the transformer and CNN branches, respectively, focus on global context and long-range dependencies and fine-grained texture and shape cues robust to quality degradation.

\subsubsection{Teacher Network: Swin-Tiny Transformer Branch}

The teacher network is constructed using the Swin-Tiny Transformer, a hierarchical vision Transformer that operates on non-overlapping patch embeddings and passes them through multiple Swin-Tiny Transformer stages. An input image $x \in \mathbb{R}^{224 \times 224 \times 3}$, prior to being linearly projected into token embeddings, the image is first divided into patches:
\begin{equation}
	\label{eq:token}
	z_i^{(0)} = W_p p_i + b_p,
\end{equation}
where $p_i$ is the $i$-th image patch, $W_p$ is the linear projection weight, and $b_p$ is the bias. These tokens go through $L$ Swin-Tiny Transformer stages, where windowed self-attention and shifted-window attention are applied at each stage to obtain multi-scale and hierarchical global context:
\begin{equation}
	\label{eq:swinblock}
	z^{(l+1)} = \text{Swin-Tiny Block}\big(z^{(l)}\big), \quad l = 0, \dots, L-1.
\end{equation}
The teacher feature vector is obtained by Global Average Pooling (GAP) over the outputs of the last stage:
\begin{equation}
	\label{eq:swinblock}
	F_T = \text{GAP}\big(z^{(L)}\big) \in \mathbb{R}^{d_T},
\end{equation}
which generates teacher logits after being run via a linear classification head:
\begin{equation}
	\label{eq:teacherlogit}
	z_T = W_T F_T + b_T,
\end{equation}
where $W_T$ and $b_T$ are, respectively, the weights and bias of the classification head.

\subsubsection{Student Network: ResNet50 Branch}

The student network is constructed using ResNet50. It is a 50-layer residual CNN that excels at extracting local texture and shape attributes via hierarchical convolutions. The student network captures a feature vector via
\begin{equation}
	\label{eq:studentfeature}
	F_S = S(x), \quad F_S \in \mathbb{R}^{d_S},
\end{equation}
after global average pooling over the last residual block. As the CNN and Transformer feature spaces are different ($d_S \neq d_T$), to harmonize heterogeneous feature spaces, we present a learnable feature adaptor $A(\cdot)$ which projects CNN features into the transformer feature space for semantic alignment: 
\begin{equation}
	\label{eq:featurespace}
	F_A = A(F_S) = W_A F_S + b_A, \quad F_A \in \mathbb{R}^{d_T},
\end{equation}

where $F_S \in \mathbb{R}^{d_S}$, $d_S$,$F_T \mathbb{R}^{d_T} $, $d_T$,$F_A \in \mathbb{R}^{d_T}$, $W_A \in \mathbb{R}^{d_T}$, and $b_A \in \mathbb{R}^{d_T}$ represent, respectively, feature vector from student branch, dimension of student feature vector, feature vector from teacher branch, dimension of teacher feature vector, adapted student feature mapped to teacher space, learnable linear projection matrix ($d_T \times d_S$), and learnable bias vector ($d_T$). This alignment enables stable joint optimization as well as meaningful interaction between the two streams. The student logits are further given by:
\begin{equation}
	\label{eq:studentlogit}
	z_S = W_S F_A + b_S,
\end{equation}
where $W_S$ and $b_S$, respectively, are the parameters of the student classification head.

\begin{figure*}[t!]
	\centering
	\includegraphics[width=0.97\textwidth,
	height=0.85\textheight,
	keepaspectratio]{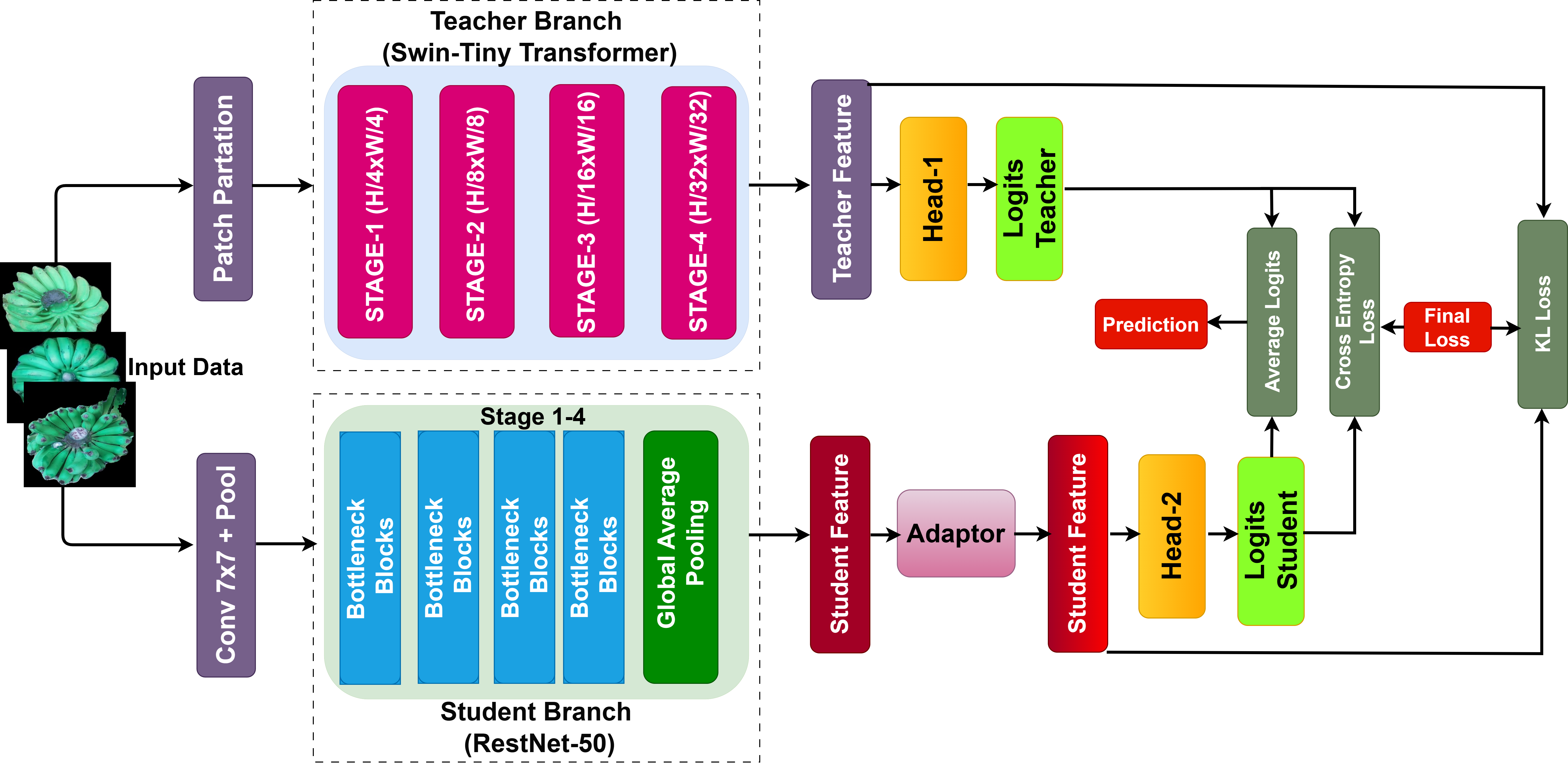}
	\caption{
		Illustration of the KL Divergence-based distillation mechanism used to align feature distributions. 
	}
	\label{fig:kl_loss}
\end{figure*}

Figure \ref{fig:kl_loss} depicts the overall workflow of the proposed hybrid framework. The ResNet50 student and the Swin-Tiny Transformer teacher process the input image simultaneously. Their outputs are fused to generate the final prediction. During training, an auxiliary KL divergence path ensures. The figure illustrates how self-distillation and supervised learning work together within the framework.

\subsubsection{Logit-Level Fusion}
Every branch?s classification head generates logits:
\begin{equation}
	\label{eq:logitfusion}
	z_T = W_T F_T + b_T, \qquad     z_S = W_S F_A + b_S.
\end{equation}

To increase robustness, logit averaging is used:
\begin{equation}
	\label{eq:avglogit}
	z_{\text{avg}} = \frac{z_T + z_S}{2}.
\end{equation}

Softmax over the fused logits is used to obtain the final prediction:
\begin{equation}
	\label{eq:logitfinal}
	\hat{y} = \arg\max \; \text{Softmax}\!\left(z_{\text{avg}}\right).
\end{equation}

\subsubsection{KL-Divergence Self-Distillation}
The main difference between the KL-distillation hybrid and the standard fusion hybrid is the addition of a feature-level KL-divergence loss that explicitly aligns the internal representations of the two branches during training. Instead of supervising only the fused output, this framework treats the Swin branch as a soft teacher whose intermediate feature distribution guides the student (ResNet50) branch, thereby performing intra-model self-distillation. As shown in Figure \ref{fig:kl_loss}, both branches produce independent class logits which are averaged at inference time, while distillation operates at the feature level.

The teacher features $F_T$ are produced by the Swin backbone. The student features $F_A$ are the adapted ResNet features after passing through a multi-layer adaptor with a residual projection:
\begin{equation}
	F_A = A(F_S) + W_r F_S
\end{equation}
where $A(\cdot)$ is the adaptor MLP and $W_r$ is the residual projection. Temperature scaling $T$ is applied to these feature vectors to produce soft distributions for the KL divergence:
\begin{equation}
	\label{eq:softmaxKL}
	p_T^{(T)} = \text{Softmax}\!\left(\frac{F_T}{T}\right), \qquad
	p_A^{(T)} = \text{LogSoftmax}\!\left(\frac{F_A}{T}\right).
\end{equation}
The KL-divergence loss then measures how well the student feature distribution matches the teacher's, scaled by $T^2$ to compensate for the gradient reduction introduced by temperature scaling:
\begin{equation}
	\mathcal{L}_{\text{KL}} = T^2 \cdot D_{\text{KL}}\!\left(p_A^{(T)} \,\|\, p_T^{(T)}\right)
\end{equation}
Note that $F_T$ is detached from the computation graph, so gradients from $\mathcal{L}_{\text{KL}}$ only update the student branch, preventing the teacher from being pulled towards the student. Both branches are simultaneously supervised with cross-entropy against the ground-truth labels:
\begin{equation}
	\mathcal{L}_{\text{total}} = \mathcal{L}_{\text{CE}}(z_T,\, y) + \mathcal{L}_{\text{CE}}(z_S,\, y) + \alpha\, \mathcal{L}_{\text{KL}}
\end{equation}
where $z_T$ and $z_S$ are the teacher and student output logits respectively, $y$ is the ground-truth label, and $\alpha = 0.5$ controls the strength of the distillation signal. This $\mathcal{L}_{\text{total}}$ is the final training loss used directly for backpropagation. The complete forward pass is:
\begin{equation}
	x \;\rightarrow\;
	\left\{
	\begin{aligned}
		F_T &= \text{Swin}(x), \\
		F_S &= \text{ResNet}(x), \\
		F_A &= A(F_S) + W_r F_S, \\
		z_T &= \text{Head}_T(F_T), \quad z_S = \text{Head}_S(F_A), \\
		z_{\text{avg}} &= \dfrac{z_T + z_S}{2}, \\[6pt]
		\mathcal{L}_{\text{total}} &= \mathcal{L}_{\text{CE}}(z_T, y) + \mathcal{L}_{\text{CE}}(z_S, y) + \alpha\,\mathcal{L}_{\text{KL}}
	\end{aligned}
	\right.
\end{equation}
At inference time, $z_{\text{avg}}$ is used for the final class prediction. To assess the impact of this distillation objective, we compare two models sharing the same dual-stream architecture: the Standard hybrid, trained with cross-entropy on the concatenation-fused output only, and the KL-Distillation hybrid, trained with the inter-stream feature-level KL loss in addition to both branch CE losses. This directly isolates the benefit of distributional alignment beyond feature fusion.

\section{Experiments Setups}\label{sec:Experiments}

In this section, we present the experimental evaluation of the proposed Cross-Grade banana variety classification scheme.

\subsection{Experimental Design}

We implement a Leave-One-Grade-Out (LOGO) cross-validation strategy. Crucially, to ensure robust evaluation, \emph{models are not shared between experiments}. For every architecture, three completely independent models are instantiated and trained from scratch for the specific grade combination.
\begin{table*}[h]
\centering
\caption{Experimental Configuration for Leave-One-Grade-Out (LOGO) Cross-Validation}
\label{tab:experimental_setup}
\setlength{\tabcolsep}{6pt} % column spacing
\begin{tabular}{ccccc}
	\toprule
	\textbf{Experiment} & \textbf{Training Grades} & \textbf{Test Grade} & \textbf{Train Size} & \textbf{Test Size} \\
	\midrule
	\textbf{1} & Grade 2 + Grade 3 & Grade 1 & 19,139 & 10,412 \\
	\textbf{2} & Grade 1 + Grade 3 & Grade 2 & 18,763 & 10,830 \\
	\textbf{3} & Grade 1 + Grade 2 & Grade 3 & 19,117 & 10,436 \\
	\bottomrule
\end{tabular}
\end{table*}

\subsection{Training Configuration}
The models were trained for up to 50 epochs, with early stopping to mitigate overfitting. The early stopping criterion was configured with a patience of 7 epochs and a minimum improvement threshold ($\text{min\_delta} = 0.001$) monitored on the validation loss. Optimization was carried out using the AdamW optimizer with an initial learning rate of $1 \times 10^{-4}$. A batch size of 32 was adopted to balance computational efficiency and convergence stability. All experiments were conducted on a CUDA-enabled H100-NVL GPU to facilitate accelerated training and efficient model optimization.

%\begin{itemize}
%   \item \textbf{Maximum Epochs}: 50
%  \item \textbf{Early Stopping}: Patience of 7 epochs, minimum delta of 0.001
% \item \textbf{Optimizer}: AdamW (LR: 1×10\textsuperscript{-4})
% \item \textbf{Batch Size}: 32
% \item \textbf{Hardware}: CUDA-enabled GPU
%\end{itemize}

%\subsection{Experimental Results}
\section{Quantitative Results}
\subsection{Overall Model Comparison}

%Table~\ref{tab:model_comparison} presents the comprehensive comparison across all architecture variants.
Table~\ref{tab:model_comparison} presents a comprehensive comparison of all evaluated architecture variants, including baseline convolutional and transformer models, as well as hybrid configurations.  Among the baseline models, Swin-Tiny Transformer achieves the highest mean accuracy (0.7294) and mean F1-score (0.7118), outperforming ResNet50 and EfficientNet-B1.  Its hierarchical transformer architecture and shifted window self-attention must therefore be responsible for its stronger capacity to model both local texture irregularities and long-range contextual dependencies. Compared to convolution-only architectures, Swin-Tiny Transformer achieves stronger discriminative representations.

The hybrid Swin-ResNet (Standard - NoAug) model further improves performance, achieving a mean accuracy of 0.7746 and a mean F1-score of 0.7443. ResNet enhances local texture sensitivity, enabling precise detection of bruise boundaries and fungal edges, while Swin-Tiny Transformer provides global contextual modeling to capture ripening patterns and spatial irregularities. The fusion of local inductive bias with long-range attention increases feature diversity and improves inter-class separability, thereby improving classification performance.

Notably, the KL divergence-based hybrid variant, Swin-ResNet (KL), achieves the best overall performance with a mean accuracy of 0.8125 and a mean F1-score of 0.8035. It also records the highest best-case accuracy (0.9209) and maintains strong worst-case performance (0.7386), demonstrating improved robustness across runs. In contrast, the Swin-EfficientNet (KL) variant does not achieve comparable gains, performing similarly to its baseline. 
Overall, the results highlight the effectiveness of hybrid architectures, particularly when enhanced with KL divergence-based regularization enforces semantic consistency, reduces feature redundancy, and improves probability calibration. This enhances inter-class separability and minority-class recall, resulting in a more robust, balanced classification.

%KL regularization enforces semantic consistency, reduces feature redundancy, and improves probability calibration. This enhances inter-class separability and minority-class recall, resulting in more robust and balanced classification.
\begin{table*}[h]
\scriptsize
\centering
\caption{Comprehensive Performance Evaluation of CNN-Based Backbones and Hybrid Architectures under KL Divergence Settings (No Augmentation)}
\label{tab:model_comparison}
\begin{tabular}{lccccc}
	\toprule
	\textbf{Model}&\textbf{Mean Acc.}&\textbf{Std Acc.}&\textbf{Mean F1}& \textbf{Best}&\textbf{Worst} \\
	\midrule
	\multicolumn{6}{l}{\textit{Baselines}} \\
	ResNet50          & 0.6945 & $\pm$0.0710 & 0.6597 & 0.7760 & 0.6465 \\
	EfficientNet-B1   & 0.7151 & $\pm$0.0390 & 0.6837 & 0.7602 & 0.6922 \\
	ConvNeXt-Tiny     & 0.7413 & $\pm$0.1155 & 0.6999 & 0.8711 & 0.6498 \\
	DenseNet-121      & 0.7858 & $\pm$0.0536 & 0.7752 & 0.8362 & 0.7295 \\
	ViT-B/16          & 0.7581 & $\pm$0.0940 & 0.7457 & 0.8338 & 0.6528 \\
	Swin-Tiny         & 0.7294 & $\pm$0.0664 & 0.7118 & 0.7980 & 0.6654 \\
	\midrule
	\multicolumn{6}{l}{\textit{Hybrids (KL Divergence - NoAug)}} \\
	\textbf{Swin-ResNet (KL)}      & \textbf{0.8125} & \textbf{$\pm$0.0946} & \textbf{0.8035} & \textbf{0.9209} & \textbf{0.7386} \\
	Swin-EfficientNet(KL)         & 0.7150 & $\pm$0.0843 & 0.6850 & 0.8044 & 0.6370 \\
	Swin-ConvNeXt(KL)             & 0.7823 & $\pm$0.1397 & 0.7543 & 0.9077 & 0.6317 \\
	Swin-DenseNet(KL)             & 0.7587 & $\pm$0.1441 & 0.7165 & 0.9168 & 0.6347 \\
	ViT-B/16--ResNet(KL)          & 0.7711 & $\pm$0.0595 & 0.7598 & 0.8257 & 0.7076 \\
	ViT-B/16--ConvNeXt(KL)        & 0.7623 & $\pm$0.1283 & 0.7212 & 0.8931 & 0.6366 \\
	ViT-B/16--EfficientNet(KL)    & 0.7391 & $\pm$0.0829 & 0.7116 & 0.8135 & 0.6498 \\
	ViT-B/16--DenseNet(KL)        & 0.7956 & $\pm$0.0974 & 0.7850 & 0.8666 & 0.6846 \\
	\bottomrule
\end{tabular}
\end{table*}

Table~\ref{tab:model_comparison} further highlights the relative performance gains achieved by the proposed hybrid variants. Compared to the strongest baseline, Swin-Tiny (mean accuracy: 0.7294), the Swin-ResNet (KL) model achieves an absolute improvement of 0.0831, corresponding to a relative accuracy gain of approximately 11.4\%. In terms of F1-score, the improvement from 0.7118 to 0.8035 represents a relative gain of 12.9\%. When compared to the standard hybrid Swin-ResNet (NoAug), the KL-regularized variant improves mean accuracy from 0.7746 to 0.8125, yielding a relative gain of 4.9\%. The corresponding F1-score improvement (0.7443 to 0.8035) reflects an 8.0\% relative increase, indicating that KL-based regularization contributes not only to higher accuracy but also to improved class-balanced performance. 
Relative to conventional CNN baselines, the gains are even more pronounced. Compared to ResNet50, the proposed Swin-ResNet (KL) achieves a 17.0\% relative improvement in mean accuracy and outperforms EfficientNet-B1 by approximately 13.6\%. These consistent relative gains across multiple baselines reinforce the effectiveness of hybridization combined with KL divergence-based regularization in enhancing both predictive performance and generalization robustness.
\begin{table*}[htbp]
	\centering
	\scriptsize
	\setlength{\tabcolsep}{4pt} % reduce column spacing
	\renewcommand{\arraystretch}{0.7} % reduce row height
	
	\caption{Performance Comparison Across Test Grades}
	\label{tab:grade_comparison}
	\resizebox{\textwidth}{!}{%
		\begin{tabular}{lrrrrr}
			\toprule
			\textbf{Model} & \textbf{Test Grade} & \textbf{Accuracy} & \textbf{Precision} & \textbf{Recall} & \textbf{F1-Score} \\
			\midrule
			\multirow{3}{*}{ResNet50} 
			& Grade1 & 0.6465 & 0.6511 & 0.6465 & 0.5907 \\
			& Grade2 & 0.6609 & 0.6581 & 0.6609 & 0.6143 \\
			& Grade3 & 0.7760 & 0.7893 & 0.7760 & 0.7741 \\
			\midrule
			\multirow{3}{*}{EfficientNet-B1} 
			& Grade1 & 0.6930 & 0.7332 & 0.6930 & 0.6432 \\
			& Grade2 & 0.6922 & 0.7112 & 0.6922 & 0.6539 \\
			& Grade3 & 0.7602 & 0.7682 & 0.7602 & 0.7540 \\
			\midrule
			\multirow{3}{*}{ConvNeXt-Tiny} 
			& Grade1 & 0.6498 & 0.6385 & 0.6498 & 0.5677 \\
			& Grade2 & 0.7030 & 0.8424 & 0.7030 & 0.6628 \\
			& Grade3 & 0.8711 & 0.8766 & 0.8711 & 0.8693 \\
			\midrule
			\multirow{3}{*}{DenseNet-121} 
			& Grade1 & 0.7295 & 0.7312 & 0.7295 & 0.7050 \\
			& Grade2 & 0.7915 & 0.8467 & 0.7915 & 0.7882 \\
			& Grade3 & 0.8362 & 0.8599 & 0.8362 & 0.8323 \\
			\midrule
			\multirow{3}{*}{ViT-B/16} 
			& Grade1 & 0.6528 & 0.6471 & 0.6528 & 0.6184 \\
			& Grade2 & 0.8338 & 0.8760 & 0.8338 & 0.8357 \\
			& Grade3 & 0.7876 & 0.7880 & 0.7876 & 0.7830 \\
			\midrule
			\multirow{3}{*}{Swin-Tiny} 
			& Grade1 & 0.6654 & 0.6930 & 0.6654 & 0.6176 \\
			& Grade2 & 0.7980 & 0.8300 & 0.7980 & 0.7927 \\
			& Grade3 & 0.7249 & 0.7265 & 0.7249 & 0.7252 \\
			\midrule
			\multirow{3}{*}{Swin-ResNet (Standard)} 
			& Grade1 & 0.6758 & 0.6837 & 0.6758 & 0.6167 \\
			& Grade2 & 0.7316 & 0.8222 & 0.7316 & 0.6995 \\
			& Grade3 & 0.9163 & 0.9208 & 0.9163 & 0.9166 \\
			\midrule
			\multirow{3}{*}{\textbf{Swin-ResNet (KL)}} 
			& Grade1 & \textbf{0.7386} & \textbf{0.7462} & \textbf{0.7386} & \textbf{0.7243} \\
			& Grade2 & 0.7781 & 0.8451 & 0.7781 & 0.7659 \\
			& Grade3 & \textbf{0.9209} & \textbf{0.9237} & \textbf{0.9209} & 0\textbf{.9203} \\
			\midrule
			\multirow{3}{*}{Swin-EfficientNet (KL)} 
			& Grade1 & 0.6370 & 0.6417 & 0.6370 & 0.5844 \\
			& Grade2 & 0.7037 & 0.8102 & 0.7037 & 0.6671 \\
			& Grade3 & 0.8044 & 0.8256 & 0.8044 & 0.8036 \\
			\midrule
			\multirow{3}{*}{Swin-ConvNeXt (KL)} 
			& Grade1 & 0.6317 & 0.5936 & 0.6317 & 0.5480 \\
			& Grade2 & 0.8074 & 0.8738 & 0.8074 & 0.8077 \\
			& Grade3 & 0.9077 & 0.9086 & 0.9077 & 0.9073 \\
			\midrule
			\multirow{3}{*}{Swin-DenseNet (KL)} 
			& Grade1 & 0.6347 & 0.6142 & 0.6347 & 0.5365 \\
			& Grade2 & 0.7247 & 0.7970 & 0.7247 & 0.6968 \\
			& Grade3 & 0.9168 & 0.9217 & 0.9168 & 0.9164 \\
			\midrule
			\multirow{3}{*}{ViT-B/16--ResNet (KL)} 
			& Grade1 & 0.7076 & 0.7097 & 0.7076 & 0.6876 \\
			& Grade2 & 0.7799 & 0.8384 & 0.7799 & 0.7670 \\
			& Grade3 & 0.8257 & 0.8283 & 0.8257 & 0.8248 \\
			\midrule
			\multirow{3}{*}{ViT-B/16--ConvNeXt (KL)} 
			& Grade1 & 0.6366 & 0.5807 & 0.6366 & 0.5231 \\
			& Grade2 & 0.7572 & 0.8539 & 0.7572 & 0.7469 \\
			& Grade3 & 0.8931 & 0.8951 & 0.8931 & 0.8937 \\
			\midrule
			\multirow{3}{*}{ViT-B/16--EfficientNet (KL)} 
			& Grade1 & 0.6498 & 0.6381 & 0.6498 & 0.5783 \\
			& Grade2 & 0.7540 & 0.8463 & 0.7540 & 0.7472 \\
			& Grade3 & 0.8135 & 0.8128 & 0.8135 & 0.8094 \\
			\midrule
			\multirow{3}{*}{ViT-B/16--DenseNet (KL)} 
			& Grade1 & 0.6846 & 0.6936 & 0.6846 & 0.6573 \\
			& Grade2 & 0.8355 & 0.8627 & 0.8355 & 0.8327 \\
			& Grade3 & 0.8666 & 0.8705 & 0.8666 & 0.8651 \\
			\bottomrule
		\end{tabular}%
	}
\end{table*}
\FloatBarrier
\subsection{Grade-Specific Performance Comparison}
Table~\ref{tab:grade_comparison} presents detailed performance metrics across three independent test grades, enabling a fine-grained evaluation of model generalization under varying data conditions. For Grade1, the KL-regularized Swin-ResNet achieves the highest performance across all evaluation metrics, attaining an accuracy of 0.7386 and an F1-score of 0.7243. This represents a notable improvement over conventional CNN baselines such as ResNet50 (0.6465 accuracy) and EfficientNet-B1 (0.6930 accuracy), indicating stronger robustness in more challenging test scenarios. In Grade2, Swin-Tiny achieves competitive performance (0.7980 accuracy), while the proposed Swin-ResNet (KL) maintains strong results with an accuracy of 0.7781 and the highest precision (0.8451), suggesting improved discriminative capability and reduced false positives. The standard Swin-ResNet hybrid also demonstrates consistent gains over individual backbones, confirming the benefit of architectural hybridization. 

Grade3 results show a clear dominance of hybrid architectures. Swin-ResNet (Standard) achieves 0.9163 accuracy, while the KL-regularized variant further improves to 0.9209 accuracy and 0.9203 F1-score, representing the best overall grade-specific performance. The consistently high recall values indicate strong class coverage and stable generalization under favorable data conditions. Across all grades, CNN-only baselines exhibit greater fluctuation in performance, particularly in Grade1 and Grade2. In contrast, the KL-based hybrid model maintains consistently strong results, demonstrating improved robustness across varying difficulty levels. These findings suggest that the integration of transformer representations with convolutional features, combined with KL divergence-based regularization, enhances both predictive accuracy and grade-wise stability.

Table~\ref{tab:grade_comparison} presents detailed performance metrics across three independent test grades, enabling a fine-grained evaluation of model generalization under varying data conditions.

For Grade1, the KL-regularized Swin-ResNet achieves the highest accuracy (0.7386) and F1-score (0.7243). Compared to the strongest baseline for this grade, EfficientNet-B1 (0.6930 accuracy), this corresponds to a relative accuracy improvement of approximately 6.6\%. Compared to ResNet50 (0.6465 accuracy), the improvement is approximately 14.2\%. The F1-score gain over the best baseline (0.6432) represents a relative improvement of about 12.6\%, highlighting enhanced class-balanced performance under more challenging conditions.

For Grade2, Swin-Tiny achieves the highest accuracy (0.7980). However, the proposed Swin-ResNet (KL) achieves 0.7781 accuracy and the highest precision (0.8451), indicating improved discriminative confidence. Compared to ResNet50 (0.6609 accuracy), the KL-based hybrid yields a relative improvement of approximately 17.7\%, and compared to EfficientNet-B1 (0.6922 accuracy), the improvement is approximately 12.4\%. These results demonstrate consistent gains over CNN-based baselines.

In Grade3, hybrid architectures clearly dominate. The Swin-ResNet (KL) achieves the best performance with 0.9209 accuracy and 0.9203 F1-score. Relative to the strongest baseline for this grade (ResNet50 at 0.7760 accuracy), this corresponds to an 18.7\% relative improvement in accuracy. Compared to Swin-Tiny (0.7249 accuracy), the gain is approximately 27.0\%, highlighting the substantial benefit of hybridization and KL-based regularization under favorable data conditions.

Overall, the relative improvements across all grades confirm that the proposed KL-regularized hybrid architecture consistently enhances predictive accuracy, class-balanced performance, and grade-wise robustness compared to both CNN-only and transformer-only baselines.

\subsection{Cross-Grade Generalization: Train-on-One-Grade Protocol} \label{sec:train_on_one}

To further assess the cross-grade generalization capability of the proposed hybrid framework, we conducted an additional evaluation under a more stringent \emph{Train-on-One-Grade} (T1G) protocol. Under this protocol, each model is trained exclusively on samples from a single quality grade and evaluated on each of the two remaining unseen grades independently. This extends the Leave-One-Grade-Out (LOGO) evaluation and examines whether the learned representations can generalize from a single source domain to multiple target domains simultaneously, capturing more extreme distribution shifts.

Table~\ref{tab:cross_grade_t1g} reports the grade-wise and aggregate performance of the proposed Swin-ResNet (KL) and the Standard Swin-ResNet (Dual CE) variants under the T1G protocol.

\begin{table*}[h]
\centering
\caption{Cross-Grade Generalization Results under Train-on-One-Grade (T1G) Protocol}
\label{tab:cross_grade_t1g}
\resizebox{\textwidth}{!}{%
	\begin{tabular}{llrrrrrr}
		\toprule
		\textbf{Model} & \textbf{Train Grade} & \textbf{Test Grade} & \textbf{Accuracy} & \textbf{Precision} & \textbf{Recall} & \textbf{F1-Score} \\
		\midrule
		\multirow{6}{*}{Swin-ResNet (KL)} 
		& Grade1 & Grade2 & 0.6044 & 0.4968 & 0.6044 & 0.5120 \\
		& Grade1 & Grade3 & 0.7756 & 0.7972 & 0.7756 & 0.7643 \\
		& Grade2 & Grade1 & 0.6177 & 0.4236 & 0.6177 & 0.4981 \\
		& Grade2 & Grade3 & 0.7220 & 0.8047 & 0.7220 & 0.6834 \\
		& Grade3 & Grade1 & 0.7108 & 0.7560 & 0.7108 & 0.6884 \\
		& Grade3 & Grade2 & 0.9148 & 0.9307 & 0.9148 & 0.9163 \\
		\midrule
		\multicolumn{3}{l}{\textbf{Mean (Swin-ResNet KL)}} & \textbf{0.7242 $\pm$ 0.1139} & -- & -- & \textbf{0.6771 $\pm$ 0.1576} \\
		\midrule
		\multirow{6}{*}{Swin-ResNet (Dual CE)} 
		& Grade1 & Grade2 & 0.5981 & 0.5084 & 0.5981 & 0.5238 \\
		& Grade1 & Grade3 & 0.7804 & 0.7857 & 0.7804 & 0.7826 \\
		& Grade2 & Grade1 & 0.6066 & 0.4124 & 0.6066 & 0.4894 \\
		& Grade2 & Grade3 & 0.7540 & 0.8089 & 0.7540 & 0.7128 \\
		& Grade3 & Grade1 & 0.6826 & 0.7128 & 0.6826 & 0.6508 \\
		& Grade3 & Grade2 & 0.8904 & 0.9104 & 0.8904 & 0.8918 \\
		\midrule
		\multicolumn{3}{l}{\textbf{Mean (Swin-ResNet Dual CE)}} & \textbf{0.7187 $\pm$ 0.1122} & -- & -- & \textbf{0.6752 $\pm$ 0.1536} \\
		\bottomrule
	\end{tabular}%
}
\end{table*}

Several observations emerge from this analysis. First, the T1G setting reveals a clear asymmetry in cross-grade transferability: models trained on Grade~3 (premium quality) generalize significantly better to unseen grades, particularly achieving the highest single-run accuracy of 0.9148 when evaluated on Grade~2. In contrast, training on Grade~1 (heavily blemished) or Grade~2 results in lower transfer accuracy, especially when tested on Grade~1 ? the most visually challenging domain. This is consistent with the intuition that high-quality training data captures clean, discriminative variety features, whereas degraded-quality images introduce noise that limits feature transferability.

Second, the Swin-ResNet (KL) model achieves a mean accuracy of 0.7242 across all six T1G transfer settings, compared to 0.7187 for the Dual CE variant, confirming that KL-divergence-based inter-stream distillation provides a consistent, albeit modest, benefit under single-source domain conditions. The corresponding F1-score improvement (0.6771 vs. 0.6752) indicates marginally more balanced class-wise predictions. Notably, both variants share the same dual-stream architecture, isolating the contribution of distributional alignment. Compared to both LOGO results, the mean accuracy under the T1G protocol is lower overall, reflecting the greater difficulty of the single-source transfer setting and the larger distribution shift between the training and test domains.

Third, all models exhibit high final training accuracy (best validation accuracy $\approx 1.000$) alongside competitive test performance, suggesting that the dual-stream architecture learns highly concentrated within-grade representations. The training-to-test generalization gap observed in the T1G protocol highlights the importance of exposure to diverse grade conditions during training, which the LOGO protocol explicitly provides. Overall, these results indicate that the proposed hybrid framework maintains robust generalization even in extreme single-source domain settings, and that KL-divergence regularization continues to provide consistent benefits across both the LOGO and T1G evaluation protocols.
\subsection{Ablation Study}
Table~\ref{tab:ablation_study_complete} presents a comprehensive ablation analysis. It highlights the individual and combined contributions of Dual CE supervision, feature alignment via the adaptor, and KL-divergence-based distillation.
\begin{table*}[h]
\centering
\caption{Ablation Study: Grade-wise and Component-wise Performance Analysis}
\label{tab:ablation_study_complete}
\resizebox{\textwidth}{!}{%
	\begin{tabular}{lcccccccc}
		\toprule
		\textbf{Ablation Study} & \textbf{G1 Acc.} & \textbf{G2 Acc.} & \textbf{G3 Acc.} & \textbf{Mean Acc. $\pm$ Std} & \textbf{Mean F1} & \textbf{Best Acc.} & \textbf{Worst Acc.} \\
		\midrule
		\multicolumn{8}{l}{\textit{Reference Baselines (Independent Training)}} \\
		Swin-Tiny (Teacher only) & 0.6654 & 0.7980 & 0.7249 & 0.7294 $\pm$ 0.0664 & 0.7118 & 0.7980 & 0.6654 \\
		ResNet50 (Student only) & 0.6465 & 0.6609 & 0.7760 & 0.6945 $\pm$ 0.0710 & 0.6597 & 0.7760 & 0.6465 \\
		\midrule
		\multicolumn{8}{l}{\textit{Ablation Variants (Isolating Components)}} \\
		w/o Adaptor (Dual CE) & 0.6370 & 0.8111 & 0.8719 & 0.7733 $\pm$ 0.1219 & 0.7577 & 0.8719 & 0.6370 \\
		Adaptor only (Single CE) & 0.6274 & 0.7507 & 0.8621 & 0.7467 $\pm$ 0.1174 & 0.7222 & 0.8621 & 0.6274 \\
		Adaptor + Dual CE & 0.6687 & 0.8700 & 0.9369 & 0.8252 $\pm$ 0.1396 & 0.8085 & 0.9369 & 0.6687 \\
		Adaptor + KL only (w/o CE) & 0.2472 & 0.1282 & 0.3750 & 0.2501 $\pm$ 0.1235 & 0.1810 & 0.3750 & 0.1282 \\
		Adaptor + Single CE + KL & 0.6292 & 0.7988 & 0.9129 & 0.7803 $\pm$ 0.1428 & 0.7497 & 0.9129 & 0.6292 \\
		\midrule
		\multicolumn{8}{l}{\textit{Proposed Full Framework}} \\
		\textbf{Swin-ResNet (Full KL)} & \textbf{0.7386} & \textbf{0.7781} & \textbf{0.9209} & \textbf{0.8125 $\pm$ 0.0946} & \textbf{0.8035} & \textbf{0.9209} & \textbf{0.7386} \\
		\bottomrule
	\end{tabular}%
}
\end{table*}

\textbf{Architectural and Objective Definitions:}
\begin{itemize}
\item \textbf{Dual Cross-Entropy (Dual CE):} Independent supervision is applied to both branches by enforcing classification loss on the teacher (Swin Transformer) and student (ResNet50) logits: $\mathcal{L}_{CE}(z_T, y) + \mathcal{L}_{CE}(z_S, y)$ This setting encourages each branch to learn discriminative representations individually..
\item \textbf{Single Cross-Entropy (Single CE):} upervision is applied only after feature fusion, i.e., on the final fused logits. This evaluates whether joint optimization alone is sufficient without explicitly guiding individual branches.
\item \textbf{KL-Divergence Distillation:} A distillation loss is introduced to align the output distributions of the student with the teacher, promoting knowledge transfer across branches. Note that ``w/o Adaptor'' variants cannot utilize KL distillation due to the dimension mismatch between Student ($d_S$) and Teacher ($d_T$) feature spaces.
\item \textbf{Full Model:} The complete architecture integrates the Adaptor module, Dual CE supervision, and KL-divergence distillation. This configuration enables both independent learning and cross-branch knowledge transfer, thereby improving robustness across varying quality grades.
\end{itemize}

\subsection{Confusion Matrix Analysis}
\begin{figure}[h]
\centering
\begin{subfigure}[b]{0.32\textwidth}
	\centering
	\includegraphics[width=\textwidth]{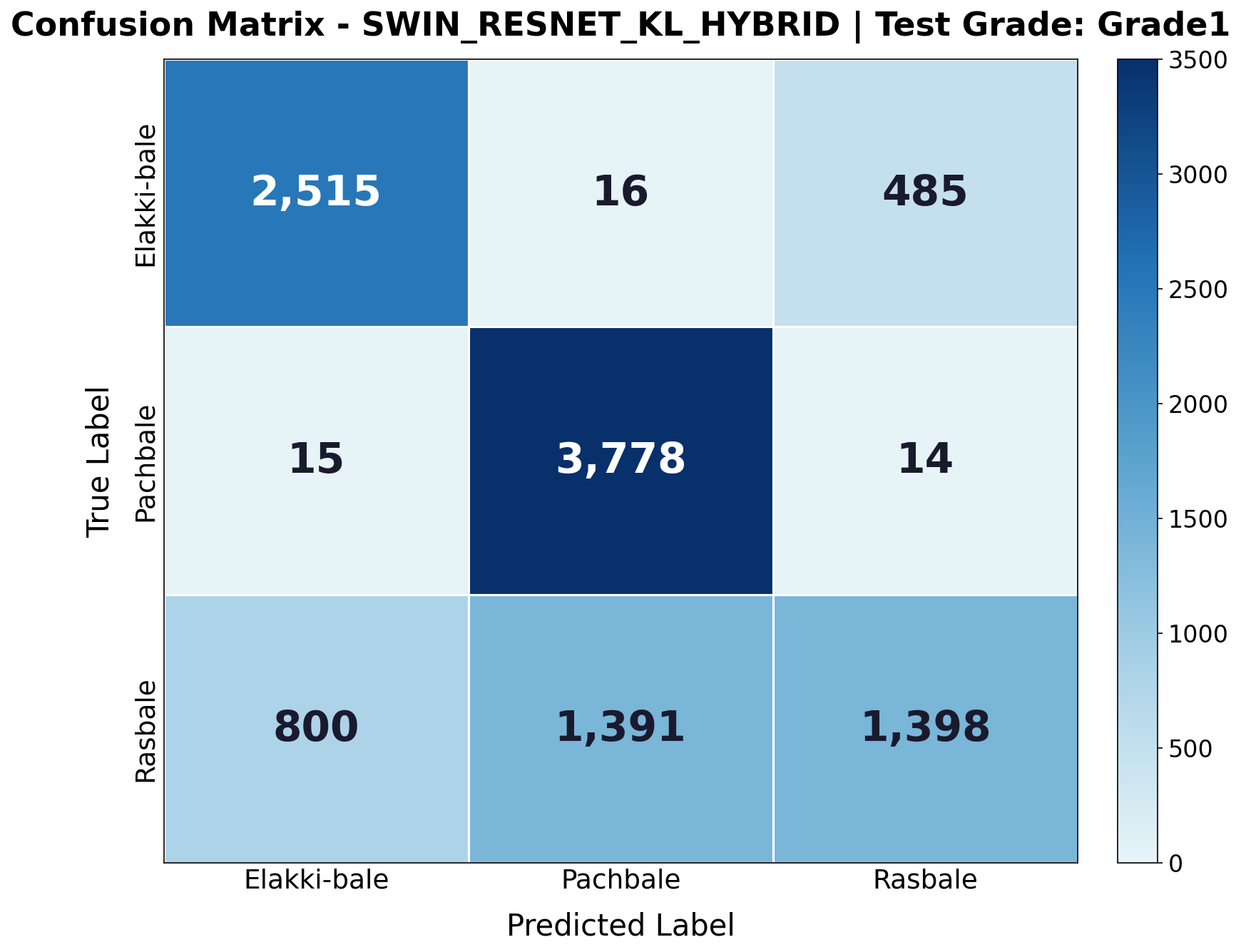}
	\caption{Grade 1 Test}
	\label{fig:kl_g1}
\end{subfigure}
\hfill
\begin{subfigure}[b]{0.32\textwidth}
	\centering
	\includegraphics[width=\textwidth]{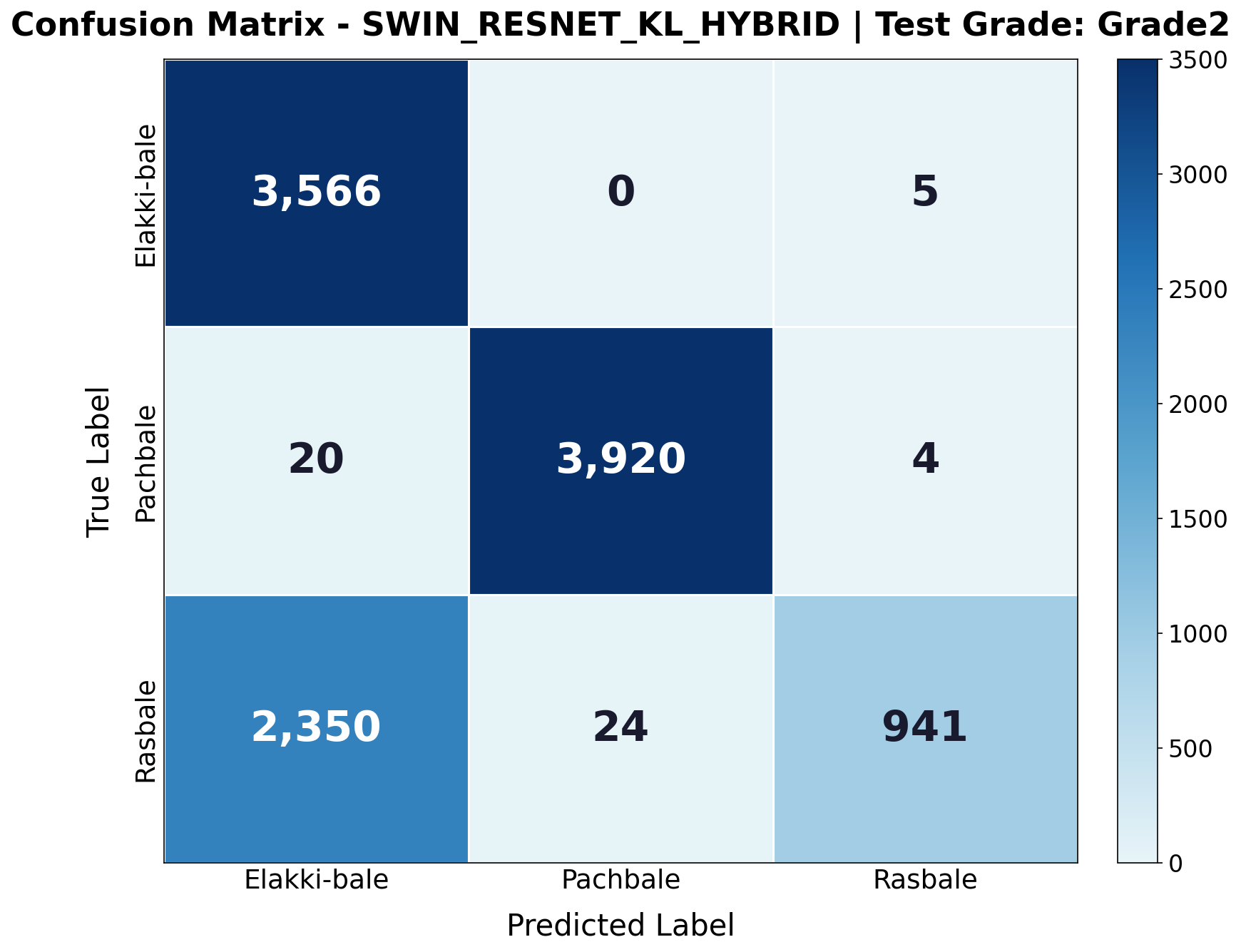}
	\caption{Grade 2 Test}
	\label{fig:kl_g2}
\end{subfigure}
\hfill
\begin{subfigure}[b]{0.32\textwidth}
	\centering
	\includegraphics[width=\textwidth]{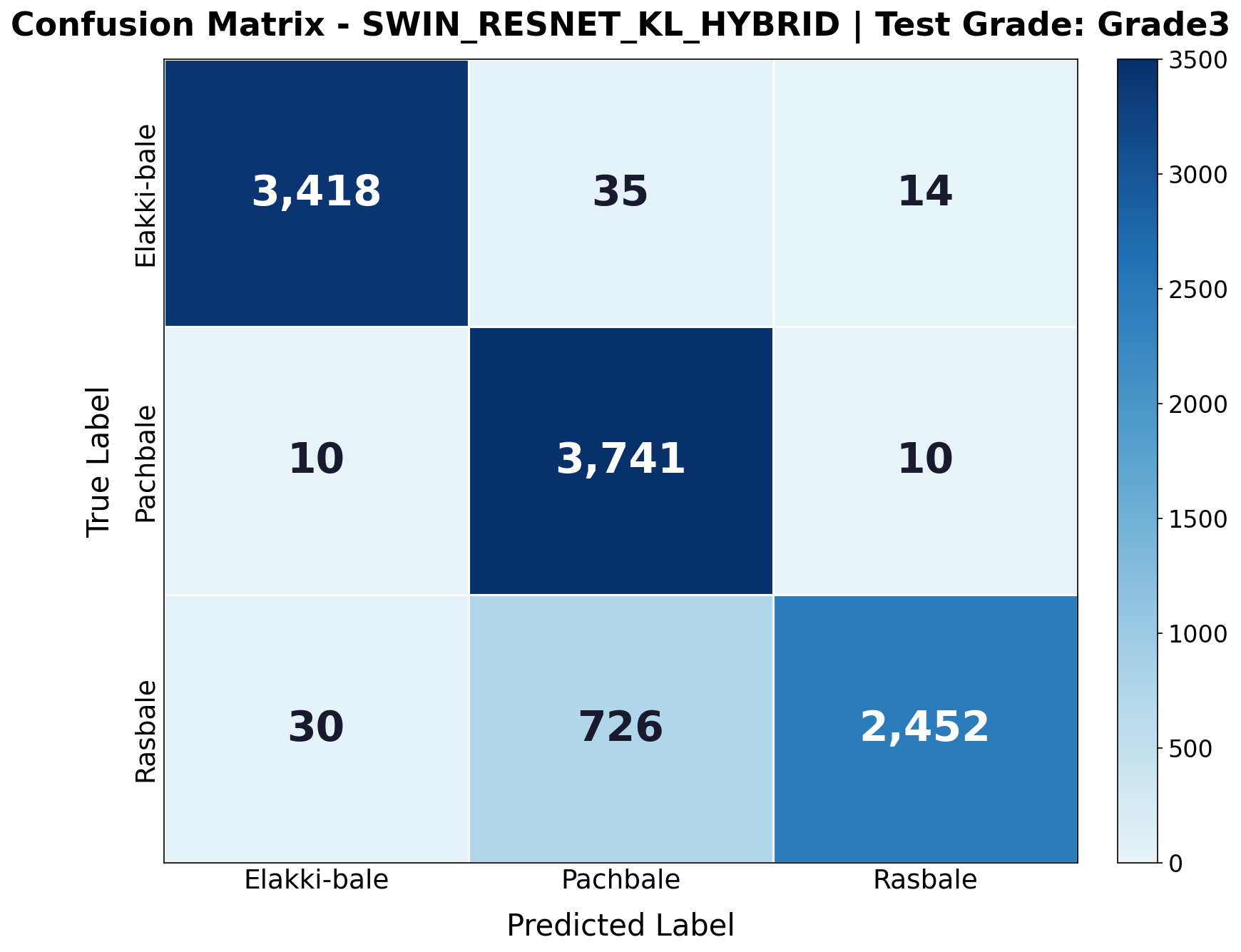}
	\caption{Grade 3 Test}
	\label{fig:kl_g3}
\end{subfigure}
\caption{Confusion Matrices for Swin-ResNet (KL) across all test grades. The strong diagonal even in Grade 1 demonstrates superior feature robustness compared to baseline CNNs.}
\label{fig:kl_matrices}
\end{figure}

Figure~\ref{fig:kl_matrices} illustrates the confusion matrices for the best-performing model, the Swin-ResNet Hybrid with KL Divergence.
The confusion matrices of the proposed Swin?ResNet (KL Hybrid) model across Grade 1 and Grade 3 demonstrate strong diagonal dominance, indicating effective class discrimination and stable cross-grade generalisation. In Grade 1, while overall classification performance remains satisfactory, relatively higher inter-class confusion is observed, particularly between Elakki-bale and Rasbale. The Rasbale class exhibits comparatively greater misclassification, suggesting higher intra-class variability at this grade level. In Grade 2, the model achieves the best performance, with near-perfect classification for Elakki-bale and minimal off-diagonal entries across all classes. This reflects enhanced class separability and well-structured feature representations. For Grade 3, high classification accuracy is maintained with limited inter-class confusion. Although minor misclassification persists for Rasbale, overall separability remains strong and superior to Grade 1.
Overall, the results confirm that the proposed KL-regularized hybrid framework learns discriminative and grade-consistent feature representations, with Grade 2 demonstrating the highest separability and Rasbale remaining the most challenging class across grades.

\subsection{Training Dynamics}
\begin{figure*}[htbp]
\centering

% --- ROW 1: GRADE 1 (Hardest Domain) ---
\begin{subfigure}[b]{0.40\textwidth}
	\centering
	\includegraphics[width=\textwidth]{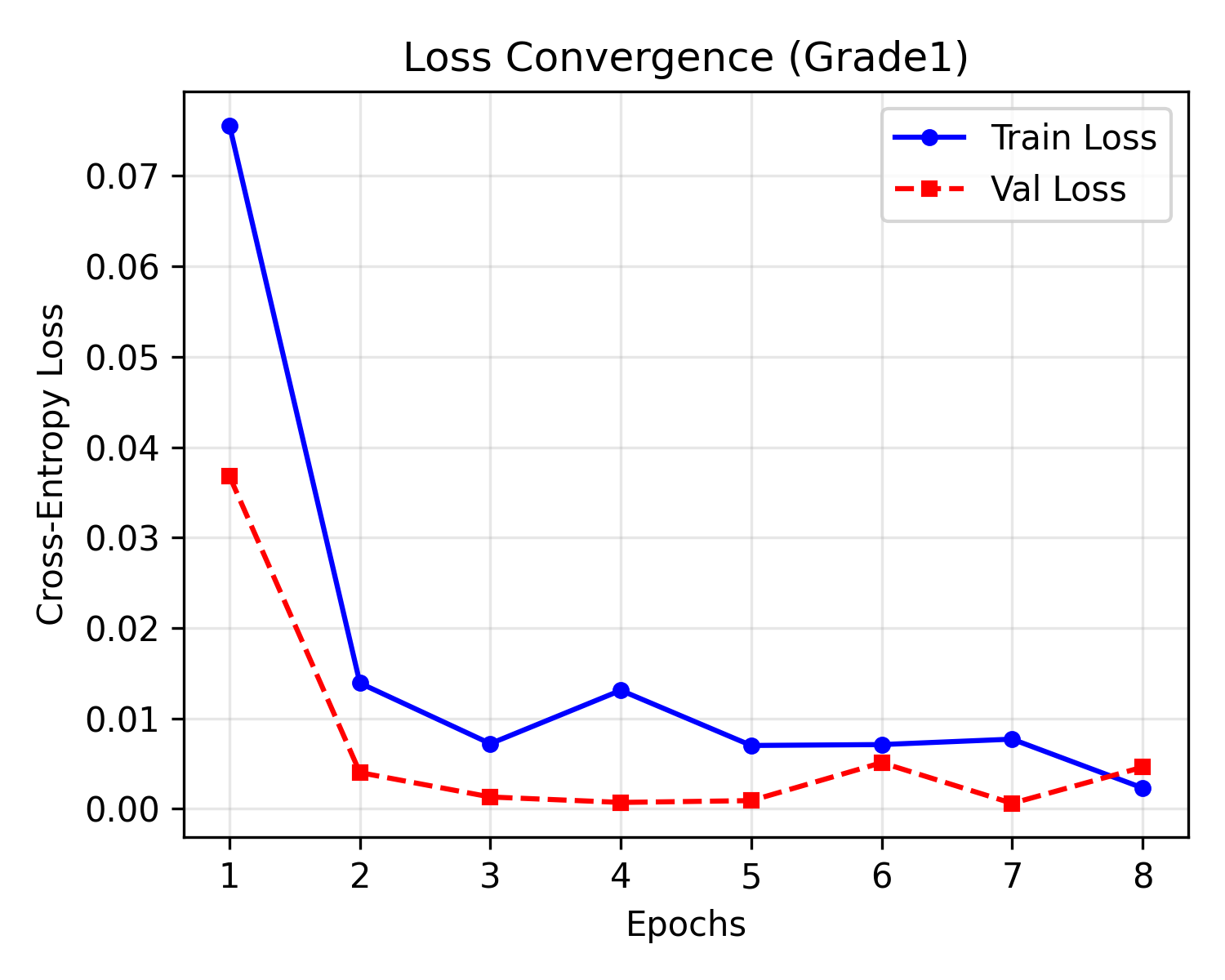}
	\caption{Loss Convergence (Test: Grade 1)}
	\label{fig:loss_g1}
\end{subfigure}
\hfill
\begin{subfigure}[b]{0.40\textwidth}
	\centering
	\includegraphics[width=\textwidth]{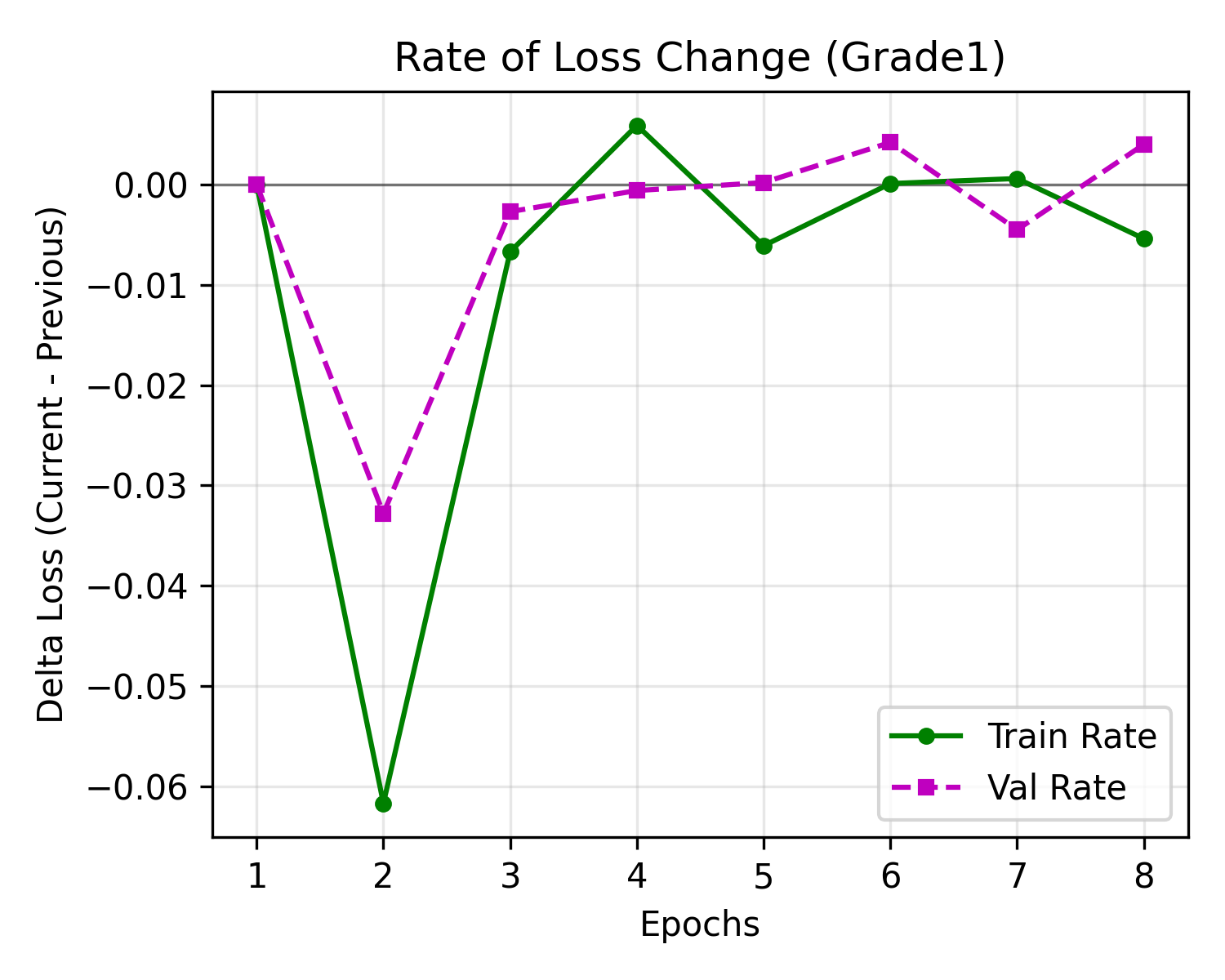}
	\caption{Rate of Change (Test: Grade 1)}
	\label{fig:rate_g1}
\end{subfigure}

%\vspace{0.2cm} % Space between rows

% --- ROW 2: GRADE 2 (Medium Domain) ---
\begin{subfigure}[b]{0.40\textwidth}
	\centering
	\includegraphics[width=\textwidth]{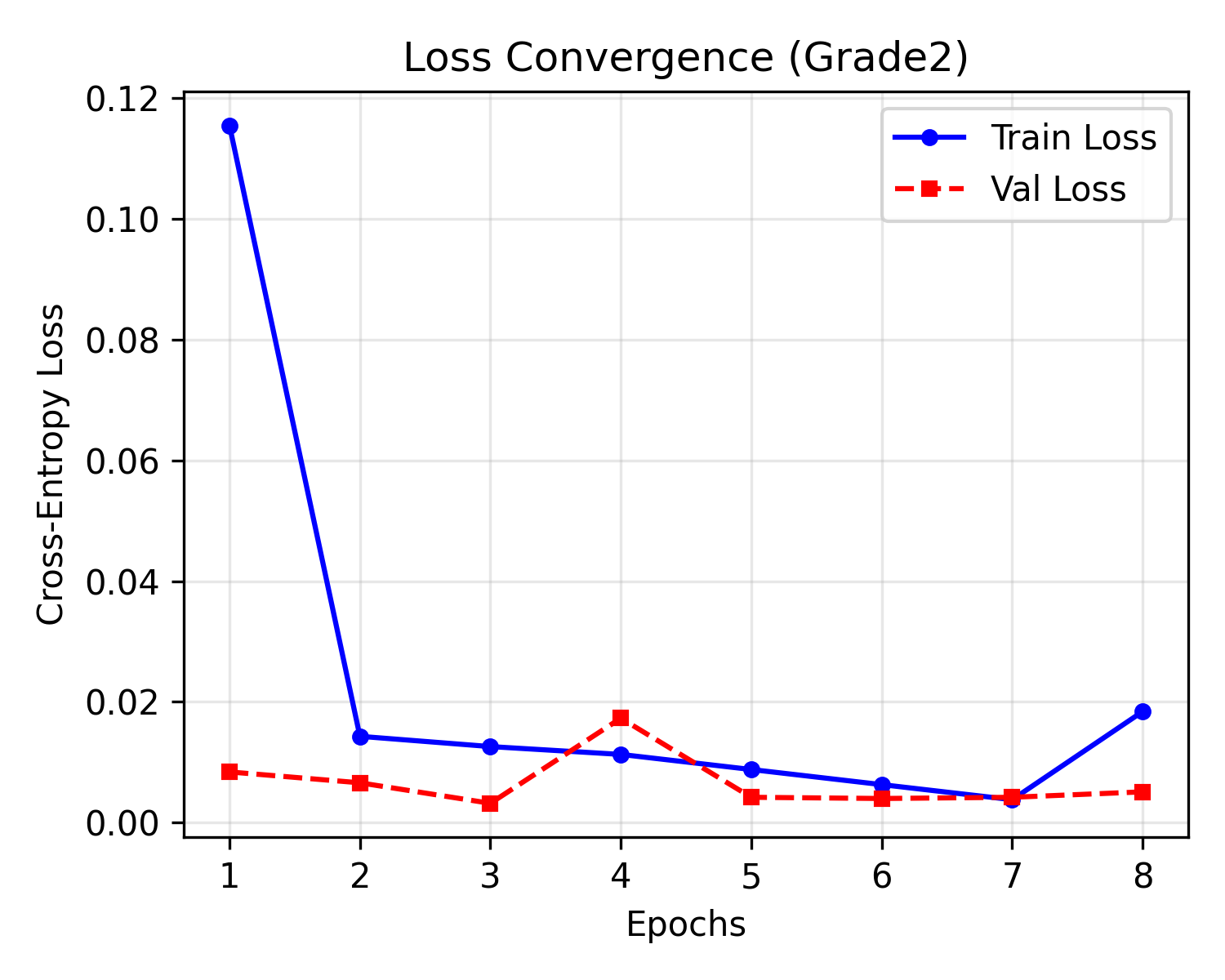}
	\caption{Loss Convergence (Test: Grade 2)}
	\label{fig:loss_g2}
\end{subfigure}
\hfill
\begin{subfigure}[b]{0.40\textwidth}
	\centering
	\includegraphics[width=\textwidth]{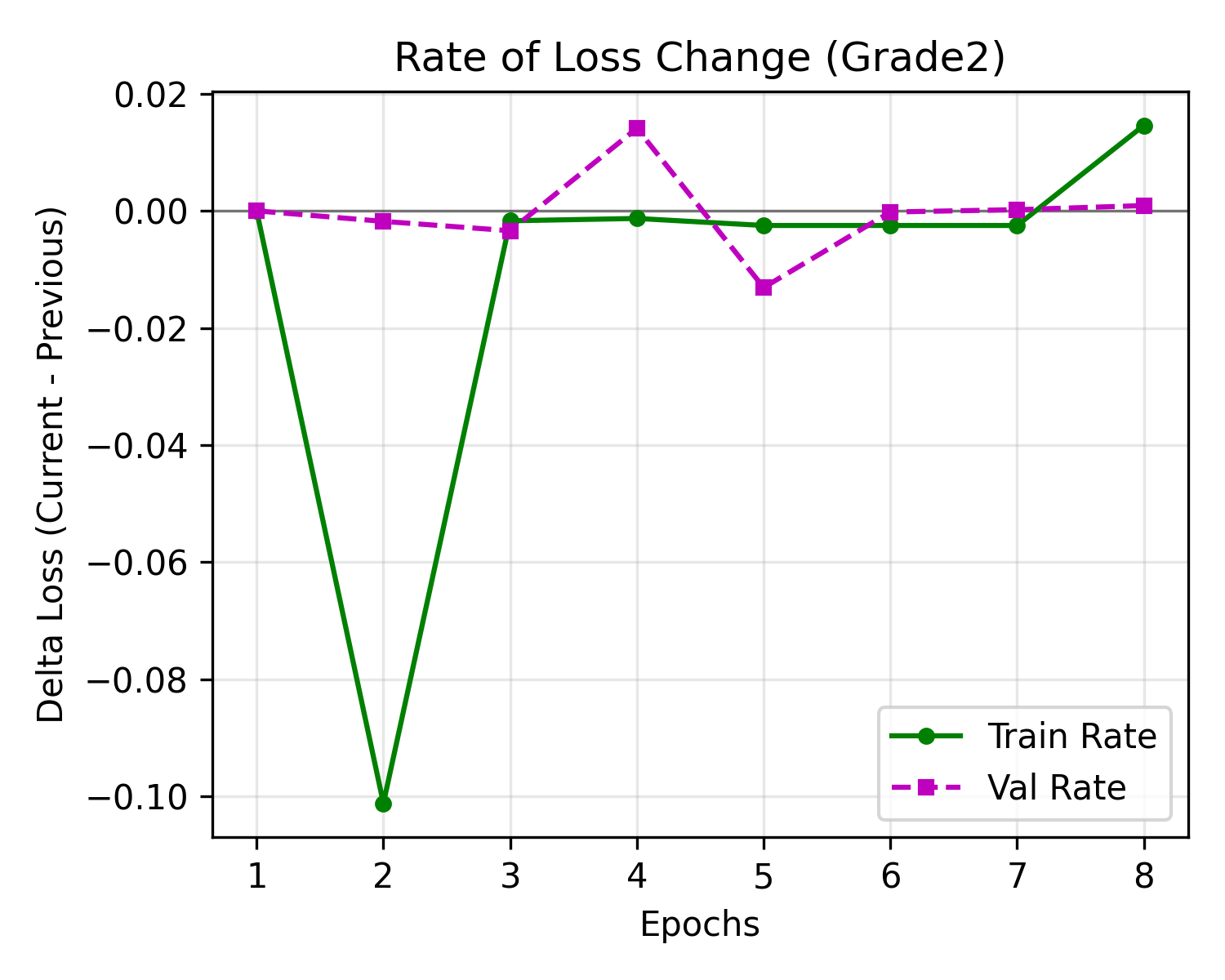}
	\caption{Rate of Change (Test: Grade 2)}
	\label{fig:rate_g2}
\end{subfigure}

%\vspace{0.2cm} % Space between rows

% --- ROW 3: GRADE 3 (Easiest Domain) ---
\begin{subfigure}[b]{0.40\textwidth}
	\centering
	\includegraphics[width=\textwidth]{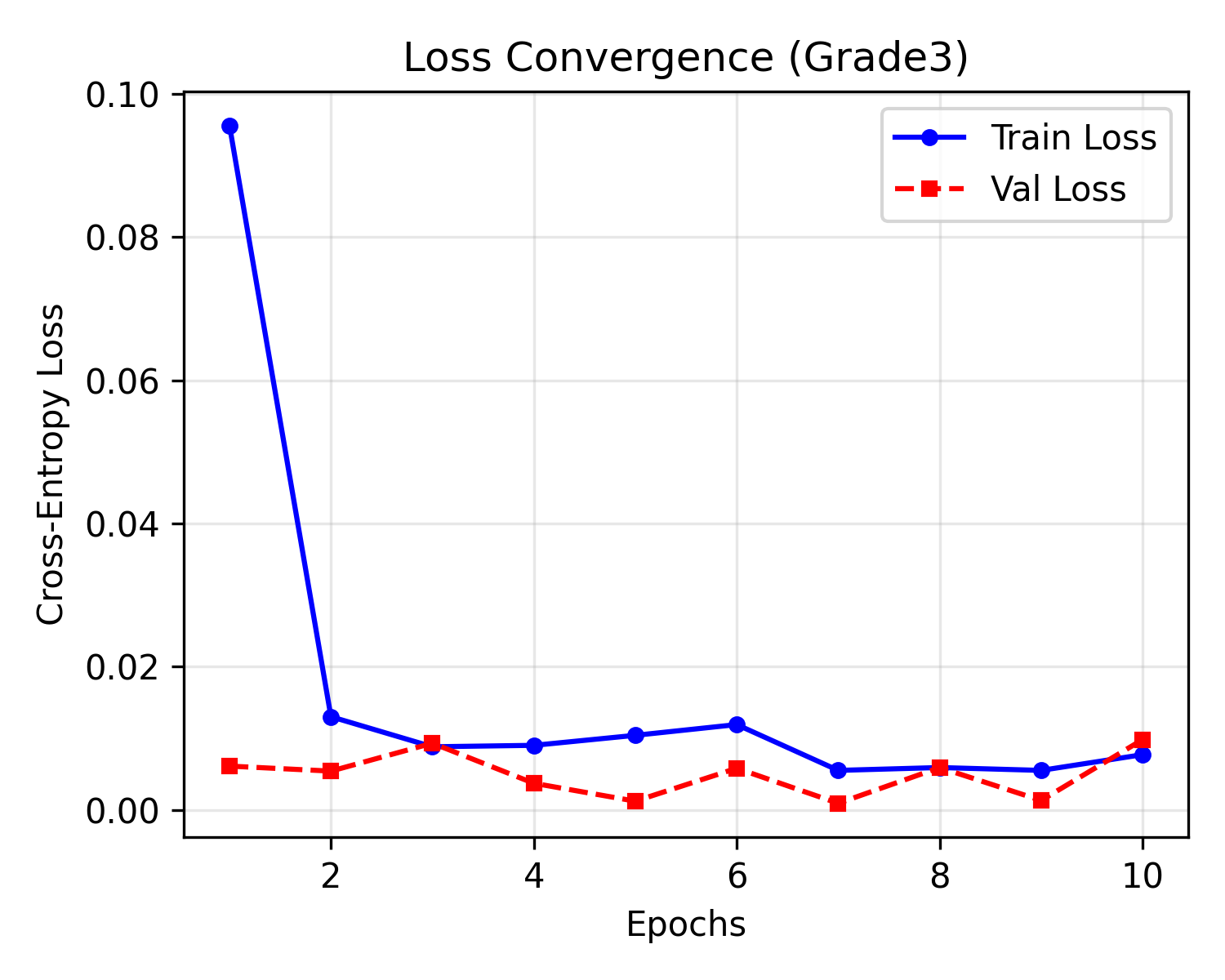}
	\caption{Loss Convergence (Test: Grade 3)}
	\label{fig:loss_g3}
\end{subfigure}
\hfill
\begin{subfigure}[b]{0.40\textwidth}
	\centering
	\includegraphics[width=\textwidth]{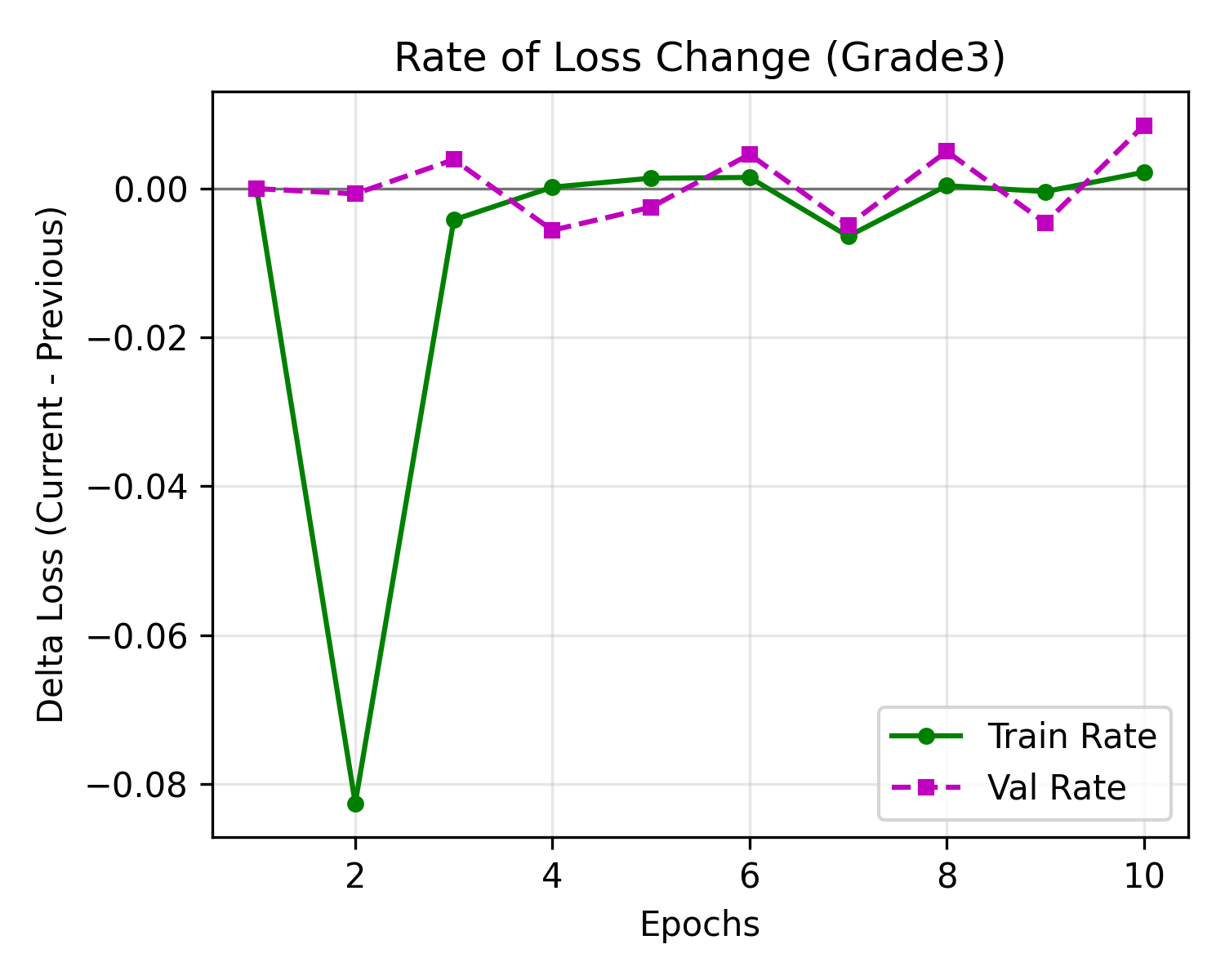}
	\caption{Rate of Change (Test: Grade 3)}
	\label{fig:rate_g3}
\end{subfigure}

\caption{\textbf{Training Dynamics Matrix.} The left column shows the cross-entropy loss profiles, where the close alignment of training (blue) and validation (red) curves indicates minimal overfitting. The right column illustrates the rate of loss change, where the rapid settling near zero ($\Delta L \approx 0$) confirms stable convergence without significant oscillations, even in the challenging Grade 1 scenario.}
\label{fig:training_dynamics_matrix}
\end{figure*}

The training stability of the Swin-ResNet Hybrid (KL Divergence) architecture was rigorously analyzed across all three Leave-One-Grade-Out (LOGO) cross-validation folds. Figure \ref{fig:training_dynamics_matrix} presents a comprehensive 3$\times$2 matrix of the training dynamics. The Left Column displays the standard Loss Convergence (Training vs. Validation), illustrating how the model minimizes error over epochs.  The consistently small gap between training and validation loss reflects good generalization, minimal overfitting, and stable convergence behavior. Overall, the model demonstrates smooth and well-controlled optimization without signs of divergence or instability.  The Right Column presents the \textit{Rate of Loss Change} ($\Delta L = L_t - L_{t-1}$), which serves as a proxy for optimization stability. The sharp negative drop in early epochs reflects strong gradient-driven updates, whereas near-zero fluctuations in later epochs confirm stabilization and fine-tuning near a local minimum.

\subsection{Training Efficiency Analysis}
Table \ref{tab:training_epochs} shows epochs to early stopping. The fewer epochs signify faster convergence. The baselines that need more training are ResNet50 (25?19?12 epochs), EfficientNet-B1 (19?18?17 epochs), and Swin-Tiny (21?24?16 epochs). The standard Swin-ResNet reduces epochs but stays slower on Grade 2. The hybrid models show clear efficiency gains. The best performing hybrids are KL-based ones, i.e.,  Swin-EfficientNet (KL) maintains its efficiency over time, and Swin-ResNet (KL) converges the quickest (8?8?10 epochs). All in all, KL-divergence alignment speeds up optimization and makes it possible for the suggested hybrids to train more quickly than independent and non-distilled models.
\begin{table*}[h]
\centering
\caption{Training Epochs Until Early Stopping}
\label{tab:training_epochs}
\begin{tabular}{lccc}
	\toprule
	\textbf{Model} & \textbf{Grade1 Test} & \textbf{Grade2 Test} & \textbf{Grade3 Test} \\
	\midrule
	ResNet50 & 25 & 19 & 12 \\
	EfficientNet-B1 & 19 & 18 & 17 \\
	Swin-Tiny & 21 & 24 & 16 \\
	Swin-ResNet(Standard) & 10 & 24 & 9 \\
	Swin-ResNet(KL) & 8 & 8 & 10 \\
	Swin-EfficientNet(KL) & 9 & 12 & 16 \\
	\bottomrule
\end{tabular}
\end{table*}

\section{Qualitative Analysis: Success vs. Failure Modes}

To better understand the model's behavior, we conducted a side-by-side analysis of successful predictions versus misclassifications across all quality grades. We specifically examine the alignment between the Student (ResNet) and Teacher (Swin) attention maps using the visualization grids generated during testing.

\subsection{Grade 1 Analysis (Hardest Domain)}

Grade 1 represents the most challenging domain due to high variability. 
\begin{figure*}[h]
\centering
\begin{subfigure}[b]{0.48\textwidth}
	\centering
	% True: Rasbale | Pred: Rasbale
	\includegraphics[width=\textwidth]{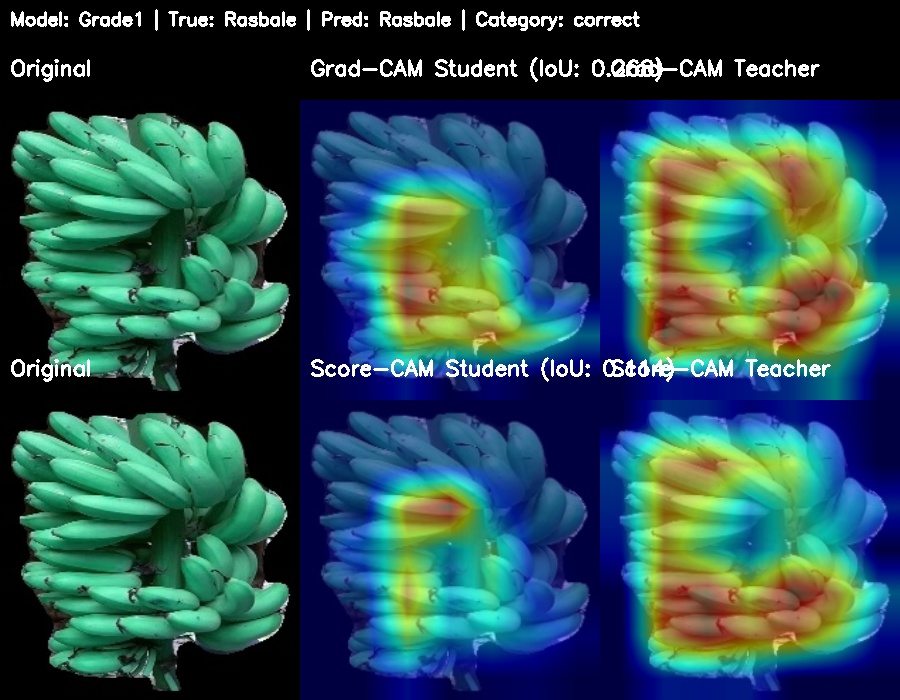}
	\caption{\textbf{Correct Prediction (Rasbale):} Strong agreement between Student and Teacher attention maps (Grad-CAM IoU: 0.268) on the fruit structure.}
	\label{fig:g1_correct}
\end{subfigure}
\hfill
\begin{subfigure}[b]{0.48\textwidth}
	\centering
	% True: Pachbale | Pred: Elakki-bale
	\includegraphics[width=\textwidth]{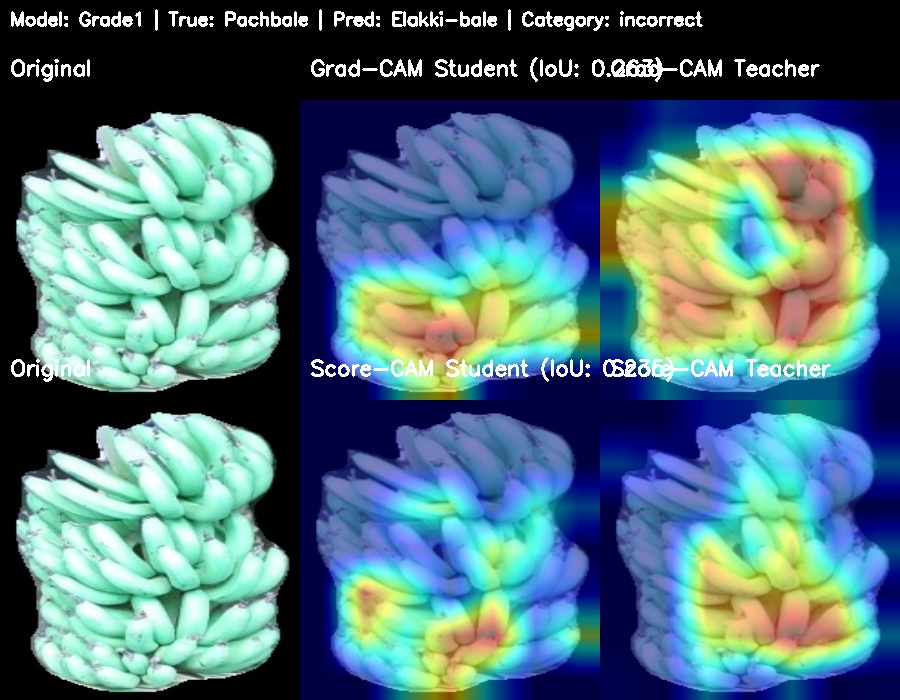}
	\caption{\textbf{Incorrect Prediction (Pachbale $\rightarrow$ Pred: Elakki):} Despite moderate attention overlap, the model confuses Pachbale with Elakki-bale, a common error in Grade1 due to similar texture profiles.}
	\label{fig:g1_incorrect}
\end{subfigure}
\caption{Grade 1 Success vs. Failure Case Study.}
\label{fig:g1_analysis}
\end{figure*}

\subsubsection{Grade 2 Analysis (Medium Quality)}
\begin{figure*}[H]
\centering
\begin{subfigure}[b]{0.48\textwidth}
	\centering
	\includegraphics[width=\textwidth]{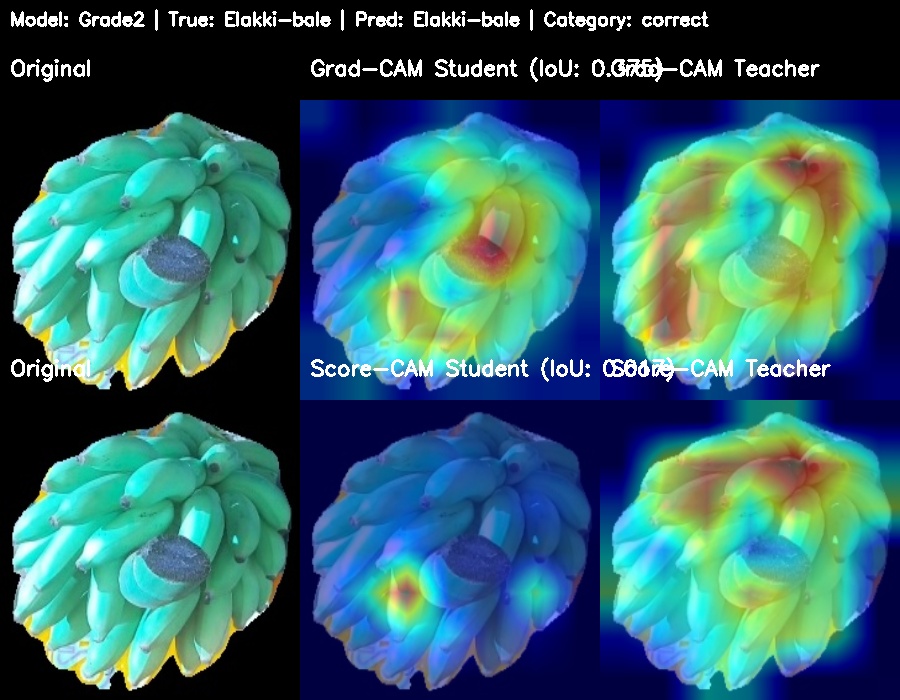}
	\caption{\textbf{Correct Classification (Elakki-bale):}    The model correctly classifies Elakki-bale, with attention concentrated on its discriminative visual features.}
	\label{fig:g2_correct}
\end{subfigure}
\hfill
\begin{subfigure}[b]{0.48\textwidth}
	\centering
	\includegraphics[width=\textwidth]{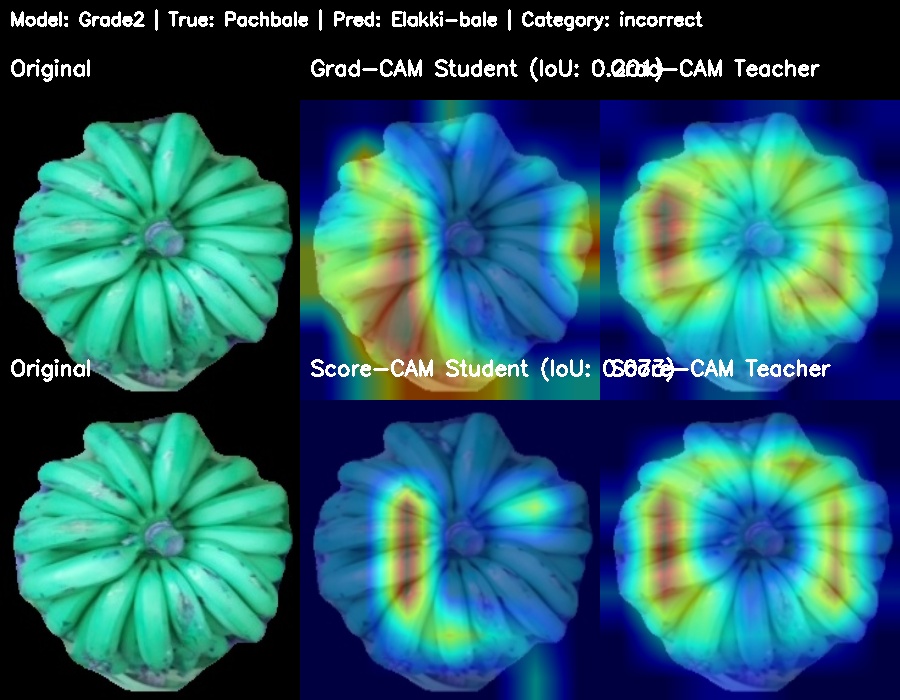}
	\caption{\textbf{Misclassification (Pachbale $\rightarrow$ Elakki-bale):} The model incorrectly classifies Elakki-bale. The attention is dispersed (low IoU), failing to capture the discriminative features of Pachbale.}
	\label{fig:g2_incorrect}
\end{subfigure}
\caption{Qualitative comparison of correct and incorrect predictions for Grade 2 samples, highlighting differences in model attention and feature localization.}
\label{fig:g2_analysis}
\end{figure*}

\subsubsection{Grade 3 Analysis (Easiest Domain)}

Grade 3 typically shows the highest accuracy, but misclassifications still occur when visual features are ambiguous.
\begin{figure*}[h]
\centering
\begin{subfigure}[b]{0.48\textwidth}
	\centering
	\includegraphics[width=\textwidth]{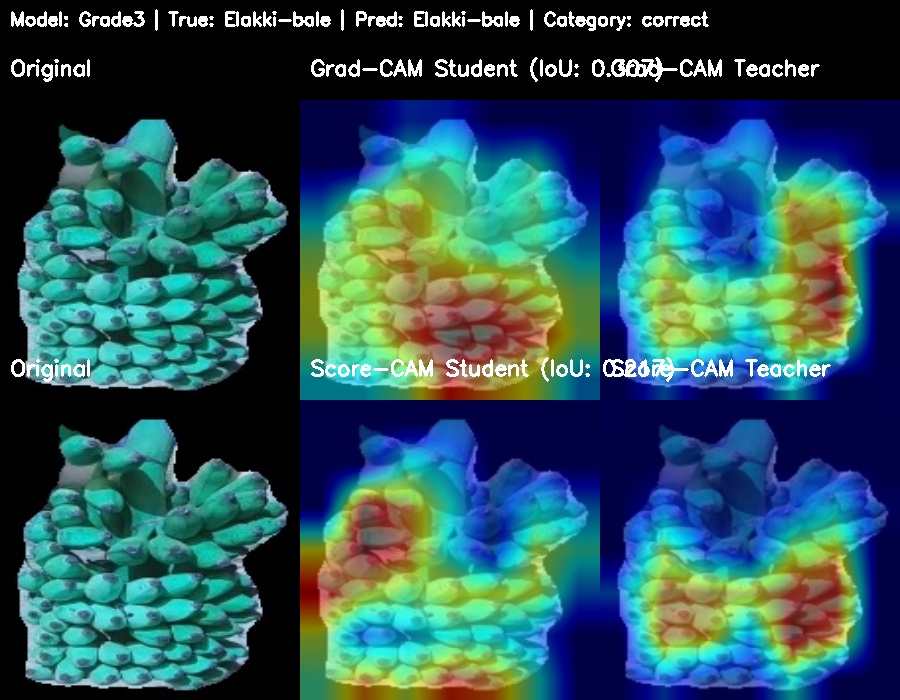}
	\caption{\textbf{Correct Classification (Pachbale):} Accurate localization of the fruit bunch, with the Student model effectively aligning with the Teacher?s attention and capturing key discriminative features.}
	\label{fig:g3_correct}
\end{subfigure}
\hfill
\begin{subfigure}[b]{0.48\textwidth}
	\centering
	\includegraphics[width=\textwidth]{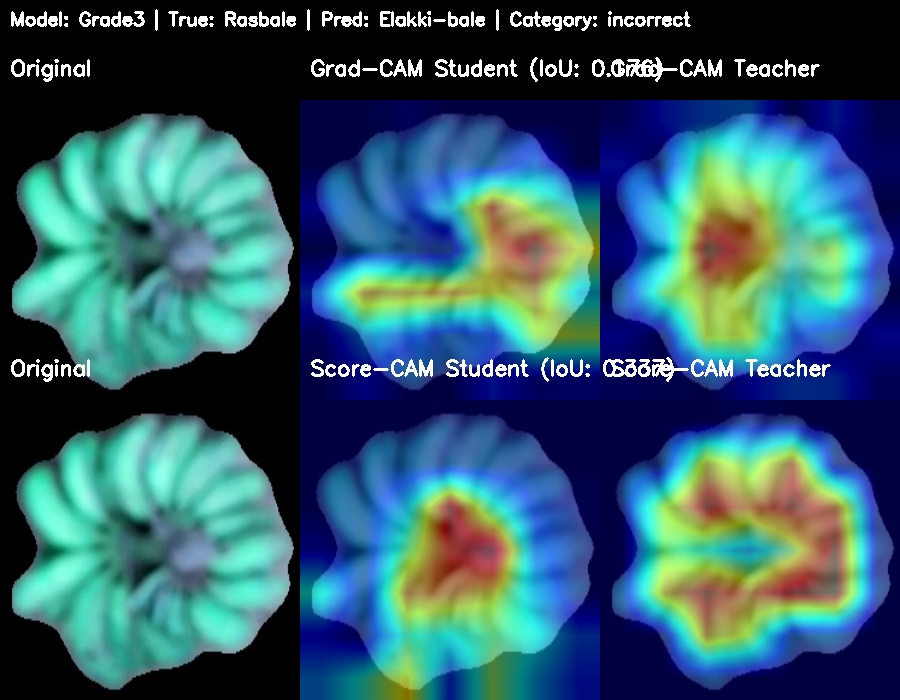}
	\caption{\textbf{Misclassification (Rasbale $\rightarrow$ Elakki-bale):} Despite focusing on the relevant region (IoU = 0.176), the model fails to distinguish fine-grained visual characteristics, leading to an incorrect prediction.}
	\label{fig:g3_incorrect}
\end{subfigure}
\caption{Qualitative comparison of correct and incorrect predictions for Grade~3 samples, illustrating the model?s localization capability and challenges in fine-grained discrimination.}
\label{fig:g3_analysis}
\end{figure*}

\subsection{Interpretability Analysis: Student vs. Teacher Attention}
\begin{figure*}[!ht]
\centering
% Wider column width since we only have 3 images per row now
\setlength{\tabcolsep}{2pt} 

% --- HEADERS ---
\begin{tabular}{ccc}
	\footnotesize \textbf{Grade 1 (Sample 1)} & \footnotesize \textbf{Grade 2 (Sample 3)} & \footnotesize \textbf{Grade 3 (Sample 4)} \\
\end{tabular}

% --- ROW 1: GRAD-CAM ---
\begin{subfigure}[b]{0.32\textwidth}
	\includegraphics[width=\textwidth]{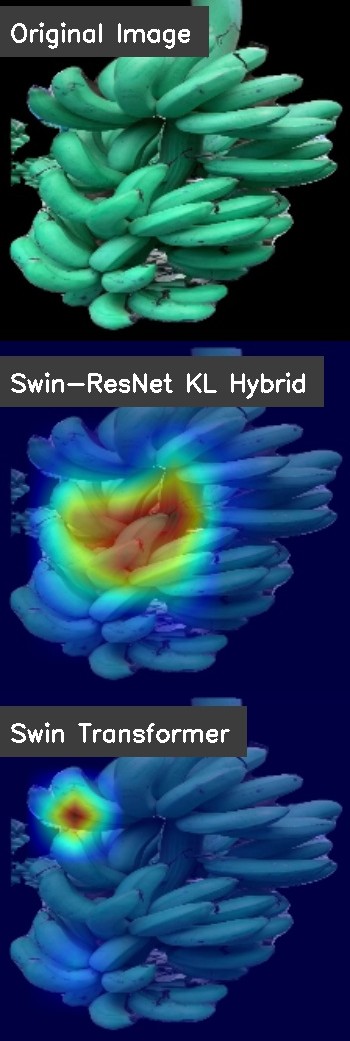}
\end{subfigure}
\hfill
\begin{subfigure}[b]{0.32\textwidth}
	\includegraphics[width=\textwidth]{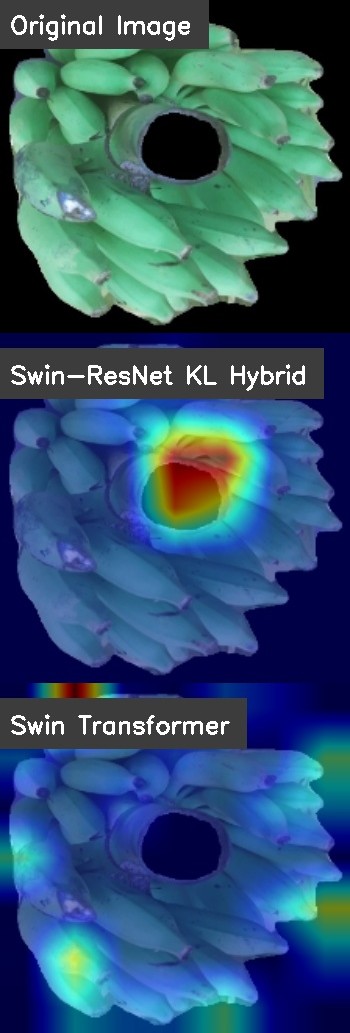}
\end{subfigure}
\hfill
\begin{subfigure}[b]{0.32\textwidth}
	\includegraphics[width=\textwidth]{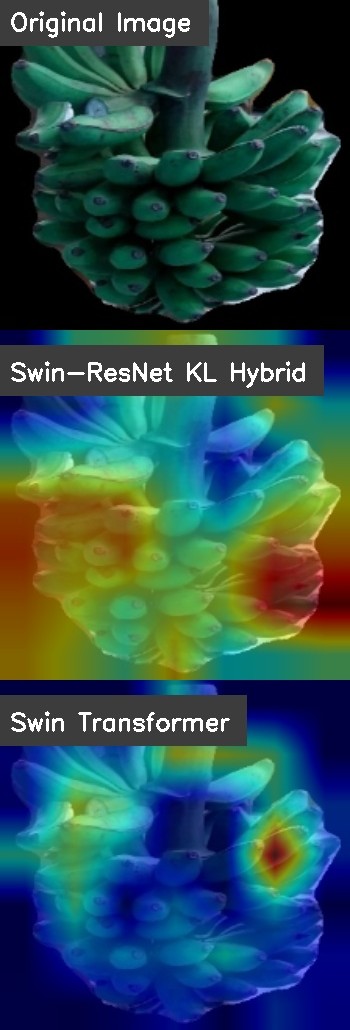}
\end{subfigure}

% \vspace{0.1cm}
\textbf{\small Row 1: Grad-CAM Comparison (Target: ResNet50 Layer4)}
\vspace{0.2cm}

% % --- ROW 2: SCORE-CAM ---
% \begin{subfigure}[b]{0.32\textwidth}
	%     \includegraphics[width=\textwidth]{Report/ScoreCAM_Comparisons/Grade1_correct_sample1_ScoreCAM_stacked.jpg}
	% \end{subfigure}
% \hfill
% \begin{subfigure}[b]{0.32\textwidth}
	%     \includegraphics[width=\textwidth]{Report/ScoreCAM_Comparisons/Grade2_correct_sample3_ScoreCAM_stacked.jpg}
	% \end{subfigure}
% \hfill
% \begin{subfigure}[b]{0.32\textwidth}
	%     \includegraphics[width=\textwidth]{Report/ScoreCAM_Comparisons/Grade3_correct_sample4_ScoreCAM_stacked.jpg}
	% \end{subfigure}

% \vspace{0.1cm}
% \textbf{\small Row 2: Score-CAM Comparison (Target: ResNet50 Layer4)}

\caption{\textbf{Student vs. Teacher Attention Matrix.} A 2$\times$3 comparison across all quality grades. In each stacked image, the middle strip represents the Student's (ResNet50) attention. The strong overlap between the Student's focus and the relevant fruit regions indicates effective feature learning from the Teacher, even though the target layer is strictly from the ResNet50 backbone.}
\label{fig:cam_matrix}
\end{figure*}

To qualitatively assess the decision-making behavior of the proposed hybrid architecture, we employed Gradient-weighted Class Activation Mapping (Grad-CAM) and Score-CAM to visualize class-discriminative regions. The analysis focuses on the Student branch (ResNet50) within the hybrid framework to examine how knowledge transferred from the Teacher (Swin Transformer) influences spatial attention patterns. For all visualizations, feature activations were extracted from the final convolutional block of the ResNet50 backbone (layer4[-1]), as this layer captures high-level semantic representations while preserving spatial localization. Figure \ref{fig:cam_matrix} presents a comparative visualization between the Student branch in the hybrid model and the baseline Swin Transformer.

Each sample is arranged in a three-row stacked format: the first row presents the original input image, the second row shows the Grad-CAM/Score-CAM attention map of the Student (ResNet50) within the hybrid architecture, and the third row displays the attention map of the baseline Swin Transformer (Teacher reference). The side-by-side comparison enables interpretability-driven evaluation of spatial focus consistency. Notably, the hybrid Student demonstrates enhanced localization of defect-specific regions (e.g., bruises, fungal lesions, and discoloration zones) with reduced background activation, indicating improved feature refinement through knowledge alignment. 

%Each image in the matrix below is a stacked visualization containing three vertical panels:
%\begin{enumerate}
%    \item \textbf{Top:} Original Input Image.
%    \item \textbf{Middle:} Attention Map of the Student (ResNet50) from the Hybrid Model.
%    \item \textbf{Bottom:} Attention Map of the Baseline Swin Transformer (Teacher Reference).
%\end{enumerate}

\begin{table*}[htbp]
\centering
\caption{Teacher-Student Attention Alignment (IoU Scores) for Correct Predictions}
\label{tab:iou_scores}
\begin{tabular}{llccc}
	\toprule
	\textbf{Grade} & \textbf{True Class} & \textbf{Sample ID} & \textbf{Grad-CAM IoU} & \textbf{Score-CAM IoU} \\
	\midrule
	Grade 1 & Rasbale & 9480 & 0.268 & 0.114 \\
	Grade 1 & Elakki-bale & 1623 & 0.313 & 0.202 \\
	Grade 2 & Rasbale & 10169 & 0.362 & 0.221 \\
	Grade 2 & Elakki-bale & 1905 & 0.375 & 0.017 \\
	Grade 3 & Elakki-bale & 2972 & 0.307 & 0.217 \\
	Grade 3 & Rasbale & 9392 & 0.176 & 0.337 \\
	\bottomrule
\end{tabular}
\end{table*}
\subsubsection{Quantitative Analysis: IoU Scores}

Table~\ref{tab:iou_scores} presents the quantitative Intersection over Union (IoU) scores between Grad-CAM and Score-CAM attention maps, measuring the spatial overlap of the generated activation masks. The IoU metric provides an objective assessment of attention consistency, where higher values indicate stronger agreement in the discriminative regions highlighted by both methods within the hybrid framework. 
Results show that the alignment between the two approaches is moderate and inconsistent. In general, Grad-CAM produces higher IoU values than Score-CAM, indicating more stable spatial localization (e.g., 0.362 vs. 0.221 for Grade 2 Rasbale and 0.375 vs. 0.017 for Grade 2 Elakki-bale). Score-CAM is more sensitive to perturbation-based attribution, as evidenced by its greater variability, which occasionally includes very little overlap. There are some exceptions, though; in Grade 3 Rasbale, Score-CAM is higher than Grad-CAM (0.337 vs. 0.176), suggesting that it can occasionally better align with class-relevant regions. Overall, the findings support the complementary nature of the two approaches in capturing distinct facets of model attention, as they show partial but inconsistent agreement.

%Table~\ref{tab:iou_scores} provides the quantitative Intersection over Union (IoU) scores, measuring the overlap between the attention masks of the two branches in the hybrid model.

\section{Conclusion}
\label{sec:conclusion}

Robust automated fruit classification frameworks must obtain reliable performance under different quality and post-harvest handling conditions. This study addressed the challenging issue of cross grades banana variety classification such that the frameworks are trained on specific quality grades and tested on unseen grades. Using a large-scale database with three commercial quality grades and three varieties, we comprehensively compared CNN, Vision Transformer, and hybrid architectures using a strict Leave-One-Grade-Out (LOGO) evaluation protocol. We presented a dual-stream Swin?ResNet hybrid system augmented with KL-divergence based self-distillation in order to overcome single-paradigm models? limitations. The presented hybrid approach learns more robust and grade-invariant representations by explicitly aligning predictive distributions and by using global context and local texture cues. According to experimental results, the hybrid framework achieved the best overall performance and the fastest convergence over baseline models.  The enhanced stability and discriminative power are further validated by grade-wise analysis and attention visualizations under difficult low-quality conditions. In spite of these encouraging outcomes, a number of areas still remains for future work such as if using domain generalization and data augmentation schemes could improve robustness even more in extreme visual variability, investigating lightweight hybrid variants might allow for real-time deployment in agricultural settings with limited resources, and  applying the framework to multiple fruits or regions datasets would aid in evaluating its wider generalization potential. All in all, this study shows that distillation guided CNN-Transformer hybridization is an efficient and intelligent approach for cross-grade agricultural image classification and inspection systems.

\section{Statement and Declarations}
This work is original and has been carried out independently by the authors. All sources of information used in this study have been properly acknowledged. The results presented in this manuscript have not been submitted to or published in any other journal or conference. The authors declare that no funding was received for this work. They have no conflicts of interest to disclose. This article does not involve any studies with human participants or animals conducted by the authors.

\section{Funding}
This research received no specific grant from any funding agency in the public, commercial, or not-for-profit sectors.
\section{Acknowledgements}
The authors thank the Department of Artificial Intelligence and the Department of Computer Science, SVNIT Surat, for computational resources on the dual NVIDIA H100 NVL GPU cluster. 
\section{Author Contributions}
Praveen Kumar Chandaliya and Khush Kataruka contributed to the conceptualization of study. Methodology was developed by Raghavendra Ramachandra, Khush Kataruka, and Praveen Kumar Chandaliya. Formal analysis was carried out by Mohammad Muzammil and Praveen Kumar Chandaliya, while investigation was performed by Zahid Akhtar and Praveen Kumar Chandaliya. Resources were provided by Praveen Kumar Chandaliya and Raghavendra Ramachandra. Data curation was conducted by Khush Kataruka and Mohammad Muzammil. Praveen Kumar Chandaliya and Zahid Akhtar prepared the original draft of the manuscript, and Mohammad Muzammil contributed to visualization. Supervision was provided by Praveen Kumar Chandaliya and Zahid Akhtar. All authors have read and agreed to the published version of the manuscript.

\section{Competing interests}

The authors declare that they have no competing interests.
\section{Availability of data and material} The dataset used in this study is publicly available through Mendeley Data and can be accessed directly at: \url{https://data.mendeley.com/datasets/dpjpdzdmsw/2} Users can download the complete dataset from the above repository. The dataset is systematically organized into folders according to \emph{banana variety}, \emph{grade}, and \emph{imaging scale}. All images and videos are properly labeled to facilitate reproducibility and further research.

%\section{Disclosure of Artificial Intelligence Use} Generative AI tools were used exclusively for language editing, grammar correction, and improving the readability of this manuscript. The research design, methodology, data collection, analysis, interpretation, conclusions, and references were independently developed by the authors. No AI tools were used to generate, manipulate, or interpret research results. The authors carefully reviewed all AI-assisted edits to ensure their correctness and consistency with the study.

% Detailed metadata and annotations are provided within the repository to support data exploration and experimental replication.

%\bibliographystyle{elsarticle-harv}
%\bibliography{cas-refs.bib}

%\end{thebibliography}
\end{document}